\pdfoutput=1

\documentclass[11pt]{article}

\usepackage[final]{acl} 

\usepackage{times}
\usepackage{latexsym}

\usepackage[T1]{fontenc}
\usepackage{soul}

\usepackage[utf8]{inputenc}

\usepackage{microtype}

\usepackage{inconsolata}
\usepackage{amsmath, amssymb}
\usepackage{enumitem}
\usepackage[most]{tcolorbox}

\usepackage{graphicx}

\usepackage[utf8]{inputenc}  
\usepackage{booktabs}        
\usepackage{multirow}
\usepackage{siunitx}
\usepackage{caption}

\title{Disambiguation-Centric Finetuning Makes Enterprise Tool-Calling LLMs More Realistic and Less Risky}

\author{First Author \\
  Affiliation / Address line 1 \\
  Affiliation / Address line 2 \\
  Affiliation / Address line 3 \\
  \texttt{email@domain} \\\And
  Second Author \\
  Affiliation / Address line 1 \\
  Affiliation / Address line 2 \\
  Affiliation / Address line 3 \\
  \texttt{email@domain} \\}

\author{
  \textbf{Ashutosh Hathidara\thanks{Equal contribution.}},
  \textbf{Julien Yu\footnotemark[1]},
  \textbf{Sebastian Schreiber}
\\
  SAP Labs
\\
  \small{
    \textbf{Correspondence:} \href{mailto:ashutosh.hathidara@sap.com}{ashutosh.hathidara@sap.com}
  }
}

\begin{document}
\maketitle

\begin{abstract}

Large language models (LLMs) are increasingly tasked with invoking enterprise APIs, yet they routinely falter when near-duplicate tools vie for the same user intent or when required arguments are left underspecified. We introduce \mbox{\textsc{DiaFORGE}} (\underline{Dia}logue \underline{F}ramework for \underline{O}rganic \underline{R}esponse \underline{G}eneration \& \underline{E}valuation), a disambiguation-centric, three-stage pipeline that (i) \emph{synthesizes} persona-driven, multi-turn dialogues in which the assistant must distinguish among highly similar tools, (ii) performs supervised \emph{fine-tuning} of open-source models with reasoning traces across 3~B--70~B parameters, and (iii) \emph{evaluates} real-world readiness via a dynamic suite that redeploys each model in a live agentic loop and reports end-to-end goal completion alongside static conversational metrics. On our dynamic benchmark \textsc{DiaBENCH}, models trained with \textsc{DiaFORGE} raise tool-invocation success by \textbf{27 pp}\footnote{“pp” = absolute percentage-point difference.} over GPT-4o and by \textbf{49 pp} over Claude-3.5-Sonnet, both under optimized prompting. To spur further research, we release an open corpus\footnote{HuggingFace url: \href{https://huggingface.co/datasets/SAP/diaforge-utc-r-0725}{https://huggingface.co/SAP/diaforge-utc-r-0725}} of \(\sim\!5{,}000\) production-grade enterprise API specifications paired with rigorously validated, disambiguation-focused dialogues, offering a practical blueprint for building reliable, enterprise-ready tool-calling agents.


\end{abstract}

\section{Introduction}\label{sec:intro}

\begin{figure}[ht]
\centering
\includegraphics[width=\columnwidth, trim = 1cm 1.4cm 0.5cm 0.5cm, clip]{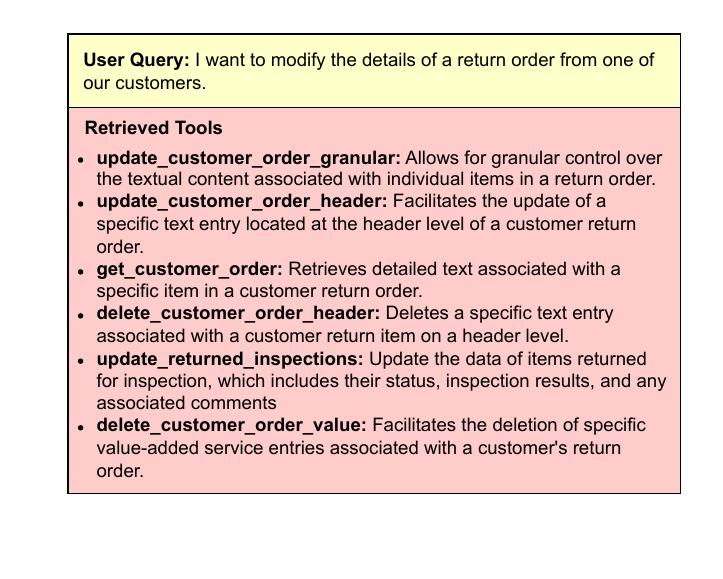}
\caption{A routine business query can retrieve multiple near-duplicate tools, illustrating the need for fine-grained disambiguation before tool invocation.}
\label{fig:near_duplicate_api_example}
\end{figure}

Modern enterprises manage \emph{thousands} of APIs, often minor variants of a core functionality customized to serve distinct domains such as customer support, finance, and supply chain operations. As LLM assistants mature from conversationalists into \emph{operational agents}, they must invoke these APIs with the same reliability that traditional software enjoys. In practice, however, single-turn user requests rarely arrive ready for direct invocation of enterprise tools: they may omit mandatory arguments, embed company-internal shorthand, or correspond to several near-duplicate tools. As Figure~\ref{fig:near_duplicate_api_example} shows, a single business query frequently surfaces multiple near-duplicate tool candidates. In our \emph{production} telemetry, \(\sim\)\textbf{35-38\%} of queries retrieve highly similar distractor APIs that require disambiguation (Appendix~\ref{sec:appendix_A4}); \(\sim\)\textbf{71\%} of live APIs declare required parameters, and \(\sim\)\textbf{76-81\%} of calls to those APIs arrive missing at least one required field. Consequently, a competent LLM assistant must master two intertwined capabilities: \textbf{multi-turn dialogue} to elicit missing arguments, and \textbf{fine-grained tool disambiguation} over a dense, overlapping API surface, often under noise and incomplete information. We address this with a disambiguation-focused pipeline for synthetic data generation and model training, empowering agents to ask targeted clarifying questions and issue accurate tool calls.


Tool-use benchmarks such as BFCL, ToolBench, and API-Bank evaluate models against \emph{fixed} user scripts, treating incoming user queries as fully specified. Each test case supplies pre-written dialogue turns, and no additional user input is generated once the assistant responds \citep{yan2024bfcl,qin2023toolllm,li2023apibank,guo2024stabletoolbench}. This off-policy setup obscures a common enterprise failure mode: under-specified requests that demand iterative back-and-forth to disambiguate near-duplicate tools and fill in missing arguments. Because static tool-use suites cannot surface the cascading-error phenomenon observed in such disambiguation-centric multi-turn exchanges \citep{laban2025llms}, our synthetic corpus intentionally withholds key details mid-dialogue and populates the tool list with semantically proximate alternatives, obliging the assistant to engage in dialogues rich in adaptive clarification. We pair model training with a dynamic evaluation harness that emulates a corporate user persona, tracking whether the model ultimately selects the correct tool and supplies required arguments. For completeness we still report static evaluation scores, but we emphasize their comparatively limited diagnostic value.

\section{Related Work}\label{sec:related_work}

\paragraph{LLMs as Tool-Using Agents.} Pioneering works such as \textsc{ReAct} interleave chain-of-thought (CoT) with tool calls, gathering evidence mid-dialogue and curb hallucinations \citep{yao2023react}. \textsc{HuggingGPT} generalizes this idea by casting LLM as a planner \citep{shen2023hugginggpt}.  These works establish language as a universal control interface for heterogeneous tools and motivate subsequent efforts to tune open models for reliable function calling.

\paragraph{Fine-Tuning LLM for Tool Use.}
\textsc{Toolformer} shows that a self-supervised annotation pipeline enables LLMs to learn when and how to invoke external utilities \citep{schick2023toolformer}.  \textsc{Gorilla} augments LLMs with API-doc retrieval, surpassing GPT-4 on tool call accuracy \citep{patil2023gorilla}.  These results imply that curated data and retrieval augmentation, not sheer parameter count, are the present keys to dependable LLM tool use.

\paragraph{Benchmarks on LLM Tool Use.}
Most widely used multi-turn benchmarks evaluate exact function call accuracy based on pre-scripted dialogues \citep{li2023apibank,yan2024bfcl,qin2023toolllm,guo2024stabletoolbench}. Recent \emph{interactive} suites broaden the evaluation scope: $\tau$-\textsc{Bench} emulates full user–agent conversations \citep{yao2024tau}; \textsc{AgentBench} spans eight environments to test long-horizon decision-making \citep{liu2024agentbench}; \textsc{MINT} and \textsc{ToolSandbox} leverage LLM-simulated user feedback \citep{wang2024mint,lu2024toolsandbox}. Most public benchmarks still overlook other enterprise-grade challenges, notably distinguishing among near-duplicate tools, proactively eliciting mandatory arguments, and detecting or preventing tool-call hallucinations, shortcomings our framework is expressly designed to remedy.

\paragraph{Data Generation and Verification.}
Verified synthetic corpora have emerged as a primary catalyst for recent gains in open-source function-calling models. \textsc{APIGen} collects thousands of executable APIs and auto-generates verified conversation traces \citep{liu2024apigen}. \textsc{ToolACE} introduces a self-evolution synthesis pipeline \citep{liu2024toolace}. \textsc{DeCRIM} employs a {decompose–critique–refine loop \citep{ferraz2024llm}. These pipelines illustrate a field-wide shift from brute-force scaling toward quality-controlled data generation driven by hierarchical feedback and automatic verification.

\paragraph{Ambiguity Resolution.}
\emph{Premature} tool invocation in response to ambiguous or underspecified requests remains an understudied failure mode for tool-augmented LLMs, especially in high-stakes enterprise settings where tool misuse can introduce significant risk. \textsc{Clarify-When-Necessary} formalizes when to ask versus act \citep{zhang2023clarify}. \textsc{CLAMBER} shows that CoT-enhanced LLMs still \emph{over-estimate} their certainty and rarely spot ambiguity \citep{zhang2024clamber}. These observations motivate our explicit disambiguation routines.

\paragraph{What Sets \textsc{DiaFORGE} Apart.}
While \textsc{APIGen}~\citep{liu2024apigen}, \textsc{ToolACE}~\citep{liu2024toolace}, and \textsc{DeCRIM}~\citep{ferraz2024llm} produce \emph{verified} function-call traces from fully specified queries, and works such as \textsc{Clarify-When-Necessary}~\citep{zhang2023clarify} theorize \emph{when} to seek clarification without training a model for it, \textsc{DiaFORGE} unifies three mutually reinforcing contributions absent from any single prior work: (i)~\emph{disambiguation-centric synthesis} that \emph{structurally} obliges the assistant to navigate near-duplicate API surfaces via injected distractors and a two-phase coercive dialogue protocol; (ii)~\emph{reasoning-trace SFT} that jointly teaches tool disambiguation and argument solicitation in a single multi-turn curriculum across 3--70~B parameter models; and (iii)~a \emph{dynamic agentic evaluation} that redeploys fine-tuned models in a live conversational loop with a simulated user, measuring end-to-end goal completion rather than isolated turn accuracy.

\begin{figure*}[t]
  \includegraphics[width=\linewidth]{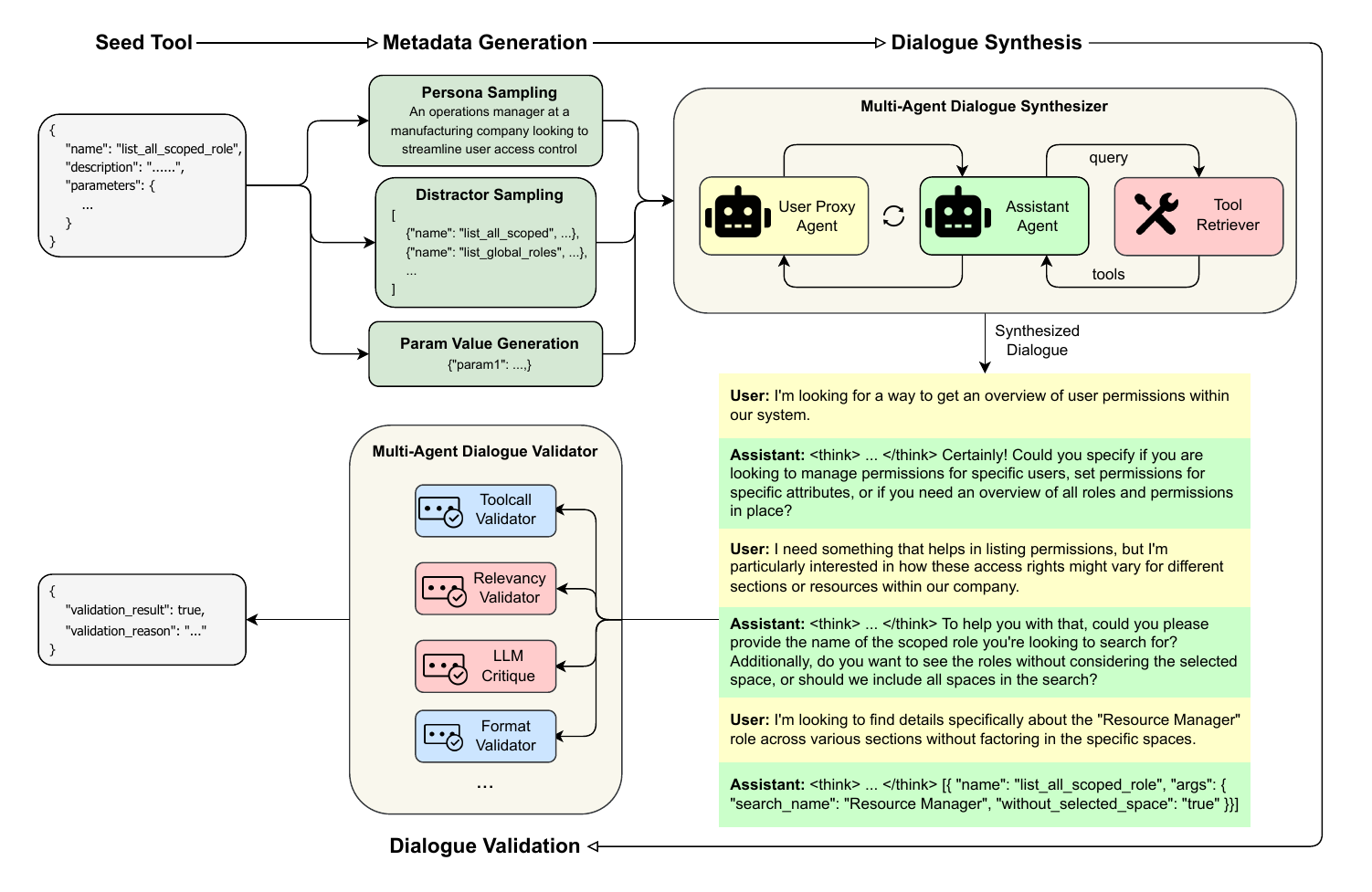}
  \caption{Data Generation Engine for Disambiguation-Centric \textbf{U}nified \textbf{T}ool-\textbf{C}alling Conversations (\textsc{UTC-Gen})}
  \label{fig:data_engine}
\end{figure*}

\section{Proposed Methodology}

Our goal is to build \emph{enterprise-grade} tool-calling LLMs that (i) accurately disambiguate near-duplicate tools and (ii) proactively request missing mandatory arguments, thereby mitigating the risk of hallucination-induced tool misuse. We present \textsc{DiaFORGE}, a three-stage pipeline encompassing synthetic dialogue generation (§\ref{sec:utc_gen}), supervised fine-tuning (§\ref{sec:sft}), and dynamic evaluation (§\ref{sec:dyna_eval}).



\subsection{Synthetic Data Generation}
\label{sec:utc_gen}

We introduce \textsc{UTC-Gen} (\underline{U}nified \underline{T}ool-\underline{C}alling \underline{Gen}erator), a multi-agent engine to construct training dialogues in a \emph{bottom-up} fashion. The engine executes three sequential phases: metadata construction, dialogue synthesis, and multi-view validation (Figure \ref{fig:data_engine}).  Each dialogue trace is \emph{seeded} with a ground-truth tool and is progressively enriched by specialized agent modules until it passes all validation gates.  Implementation details appear in Appendix~\ref{sec:appendix_A}.

\paragraph{Enterprise Tool Catalogue.}
Let
\[
  \mathcal{T}
  \;=\;
  \bigl\{
    \tau_i
    \;=\;
    \bigl(\text{name}_i,\,
          \text{description}_i,\,
          \texttt{params}_i\bigr)
  \bigr\}_{i=1}^{|\mathcal{T}|}
\]
denote the enterprise-wide set of callable tools.  For any tool $\tau_i$, the parameter specification $\texttt{params}_i$ is a JSON Schema map that associates each argument name with a triple of the form $(\texttt{type}, \texttt{description}, \texttt{required})$. We define the set of \emph{required} arguments for $\tau_i$ as $\mathcal{R}(\tau_i)$.

\paragraph{Persona Sampling.}
Given a seed tool $\tau^\star\!\in\!\mathcal{T}$, we first sample a corporate–user persona
\(
  p \;\sim\; 
  \pi^{(k)}_{\mathrm{rand}}\!\bigl(\,\cdot \mid \tau^\star,\mathcal{P}\bigr),
\)
where $\pi^{(k)}_{\mathrm{rand}}$ denotes a top-$k$ retrieval-with-randomization distribution over an enterprise-filtered subset $\mathcal{P}\subseteq\textsc{PersonaHub}$ (12 k entries) \citep{ge2024scaling}.  
Conditioned on $(\tau^\star,p)$, we instantiate a concrete user goal
\(
  g \;\sim\;
  P_{\mathrm{goal}}\!\bigl(\,\cdot \mid \tau^\star,p\bigr),
\)
which the user-proxy agent treats as its terminal objective during dialogue synthesis.

\paragraph{Distractor Tool Sampling.}
To emulate realistic tool ambiguity, we assemble a set of \emph{near-duplicate} tools.  Let $\phi:\mathcal{T}\!\to\!\mathbb{R}^{d}$ be a frozen sentence encoder applied to the concatenation of each tool’s name, description, and selected schema metadata.  We retrieve the $k=5$ semantic nearest neighbors of $\tau^\star$,
\(
  \mathcal{D}_{k}(\tau^\star)
  \;=\;
  \operatorname*{arg\,top-\textit{k}}_{\tau \in \mathcal{T} \setminus \{\tau^\star\}}
  \bigl\langle \phi(\tau^\star), \phi(\tau) \bigr\rangle.
\)
During dialogue synthesis, the assistant agent receives the candidate pool of tools
\(\mathcal{C}_k(\tau^\star) \;=\; \{\tau^\star\} \cup \mathcal{D}_k(\tau^\star),\)
and must resolve any ambiguity \emph{online} before issuing a tool call.

\paragraph{Slot Value Generator.}
We instantiate concrete, persona-consistent values for all \emph{required} slots so that the user-proxy need not invent them on the fly. Let the required arguments for \(\tau^{\star}\) be
\(
  \mathcal{R}\!\bigl(\tau^{\star}\bigr)=\{r_{1},\dots,r_{m}\}.
\)
We \emph{jointly} sample their values
\[
  \bigl(v_{r_{1}},\dots,v_{r_{m}}\bigr)\;
  \sim\;
  \mathcal{P}_{\mathrm{param}}\!\bigl(\,\cdot \mid \mathcal{R}\!\bigl(\tau^{\star}\bigr),\,p\bigr),
\]
where \(\mathcal{P}_{\mathrm{param}}\) is an LLM that, conditioned on the persona \(p\), generates realistic parameter values, such as dates, currency codes, and alphanumeric IDs, with high diversity. Aggregating the draws yields the map
\(
  \mathcal{V}^{\star}
  = \mathcal{V}\bigl(\tau^{\star},p\bigr)
  = \bigl\{(r_{i},v_{r_{i}})\bigr\}_{i=1}^{m}.
\)
During conversation simulation, the user-proxy incrementally reveals \emph{subsets} of \(\mathcal{V}^{\star}\), requiring the assistant to (i) identify disclosed values and (ii) query for any that remain unknown.

\paragraph{Dialogue Synthesis}
\label{sec:dialogue_synth}
Given a tool $\tau^\star$, persona $p$, distractor set $\mathcal{D}_{k}(\tau^\star)$, and gold argument map $\mathcal{V}^{\star}$, \textsc{UTC-Gen} synthesizes a dialogue trace \(d=\langle\,(u_{1},a_{1}),\,(u_{2},a_{2}),\dots,(u_{T},a_{T})\rangle\), where $u_{t}$ (resp.\ $a_{t}$) denotes the user (resp.\ assistant) utterance at turn~$t$.  Two running histories are maintained:
\begin{align*}
\mathbf{h}^{u}_{t}\;&=\;(u_{1},a_{1},\dots,u_{t-1},a_{t-1}),\\
\mathbf{h}^{a}_{t}\;&=\;(u_{1},a_{1},\dots,u_{t-1},a_{t-1},u_{t}),
\end{align*}
representing the context observable to the \emph{user} and \emph{assistant}, respectively, during turn~$t$.

\subparagraph{User Agent.}
At turn $t$, the user-proxy samples
\[
  u_{t}\;\sim\;
  P_{\boldsymbol{\theta}_{u}}
  \!\Bigl(
     \cdot
     \,\bigm|\,
     \tau^\star,\;
     p,\;
     g,\;
     \mathcal{D}_{k},\;
     \mathcal{V}^{\star},\;
     \mathbf{h}^{u}_{t}
  \Bigr),
\]
where $P_{\boldsymbol{\theta}_{u}}$ is the distribution induced by the user-proxy’s parameters $\boldsymbol{\theta}_u$.  The persona $p$ incorporates domain-specific jargon and tone reflective of authentic enterprise interactions, while $g$ denotes the goal of the conversation; the distractor pool $\mathcal{D}_{k}$ steers user queries toward phrasing that could match several tools, compelling the assistant to disambiguate in real time; the gold argument map $\mathcal{V}^{\star}$ bounds all slots to ground truth values, mitigating hallucination; the running history $\mathbf{h}^{u}_{t}$ preserves discourse coherence with the dialogue prefix.

The user–proxy proceeds in two successive phases, coercing the assistant to \emph{first} resolve tool ambiguity and \emph{then} guarantee slot completion:

\begin{enumerate}[label=(\roman*), leftmargin=1.65em, itemsep=2pt]
  \item \emph{Tool-selection phase.}  
        During opening turns, the user-proxy issues a deliberately under-specified request, revealing just enough context to prune the candidate set $\mathcal{C}_{k}$ until the assistant can unambiguously identify~$\tau^\star$.
  \item \emph{Argument-completion phase.}  
        After identifying $\tau^\star$, the user-proxy discloses the remaining slot values following the assistant’s requests, until every key–value pair in $\mathcal{V}^{\star}$ has been provided.
\end{enumerate}

\subparagraph{Assistant Agent.}
At turn~$t$,
\[
  a_{t}\;\sim\;
  P_{\boldsymbol{\theta}_{a}}
  \!\Bigl(
     \cdot
     \,\bigm|\,
     \mathcal{C}_{k},\;
     \mathbf{h}^{a}_{t}
  \Bigr),
\]
where \(P_{\boldsymbol{\theta}_{a}}\) is the distribution induced by the assistant LLM’s parameters \(\boldsymbol{\theta}_{a}\);  \(\mathbf{h}^{a}_{t}\) is the dialogue prefix visible to the assistant at turn~\(t\); \(\mathcal{C}_{k}\) is the set of candidate tools.  Because the assistant is \emph{oblivious} to which element of $\mathcal{C}_{k}$ is the ground-truth $\tau^{\star}$, it must  (i)~pose clarification questions that iteratively eliminate distractors, and  (ii)~solicit any missing slot values until the argument map is complete.

Each assistant turn is decomposed into a private \emph{reasoning trace} and a public \emph{response}: the former captures chain-of-thought computations internal to the model, while only the latter is revealed to the user-proxy agent.  During supervised fine-tuning (§\ref{sec:sft}), both components serve as learning targets.

\subparagraph{Stopping Criteria.}
\label{par:stop_criteria}
The simulation terminates as soon as \emph{one} of the following events occurs:
\begin{enumerate}[label=(\roman*), leftmargin=1.65em, itemsep=2pt]
  \item the assistant emits a schema-conformant call to $\tau^\star$ whose arguments map exactly to \(\mathcal{V}^{\star}\), with no missing \emph{or} superfluous keys;
  \item the dialogue length reaches the hard cap \(T_{\max}\).
\end{enumerate}

\paragraph{Validator Cascade.}
\label{par:validator}

The synthesized dialogue \(d\) enters the training corpus only if every turn stands scrutiny by a \emph{format validator}, \emph{relevancy validator}, and \emph{LLM critique} (Figure \ref{fig:data_engine}); failure at any step triggers immediate rejection (Appendix~\ref{sec:appendix_A2}).

\begin{enumerate}[label=(\alph*),leftmargin=*]
  \item \textbf{User-Utterance Validity}.  
        For every turn \(t\in\{1,\dots,T\}\), the user message \(u_t\) remains
        \begin{enumerate}[label=(\roman*),leftmargin=1.5em, nosep]
          \item coherent with the dialogue prefix \(\mathbf{h}^{u}_{t}\);
          \item grammatically intelligible and stylistically faithful to the sampled persona \(p\);
          \item semantically aligned with latent goal \(g\).
        \end{enumerate}

  \item \textbf{Assistant-Response Validity}.  
        For every turn \(t\), the assistant reply \(a_t\)
        \begin{enumerate}[label=(\roman*),leftmargin=1.5em, nosep]
          \item contains a json schema object with \emph{three} sections: a \emph{thought} trace, an optional \emph{tool\_calls} stub, and a public \emph{content};
          \item is coherent with the dialogue prefix \(\mathbf{h}^{a}_{t}\).
        \end{enumerate}
\end{enumerate}

Each dialogue must also contain one assistant turn \(t^{\dagger} \leq T\) whose \textit{tool\_calls} satisfies stopping criterion~(i). Only dialogues that pass \emph{all} validation checks are included in the final training set.

\subsection{Fine-Tuning Pipeline}
\label{sec:sft}

Let the validated corpus be
\(\mathcal{D}_{\mathrm{train}}=\{d_i\}_{i=1}^N\), with
\[
  d_i
  \;=\;
  \bigl\langle
    (u^{(i)}_{1},a^{(i)}_{1}),
    \dots,
    (u^{(i)}_{T_i},a^{(i)}_{T_i})
  \bigr\rangle.
\]
We adopt a \emph{turn-slicing} strategy \citep{ouyang2022training} (Figure~\ref{fig:data_prep_sft}): for each assistant turn \(t\in\{1,\dots,T_i\}\) we form an input–target pair
\[
  \label{eq:cum_turn_pair}
  x_{i,t}
  \;=\;
  \underbrace{\texttt{[SYS]}\;
              u^{(i)}_{1}\;
              a^{(i)}_{1}\;
              \dots\;
              u^{(i)}_{t}}_{\text{prompt context}},
  \qquad
  y_{i,t}
  \;=\;
  a^{(i)}_{t}.
\]
The model is trained \emph{only} to predict the next assistant response, given the complete dialogue prefix.

We perform standard Supervised Fine-Tuning (SFT) with LoRA \cite{hu2022lora} over next token prediction (Appendix~\ref{sec:appendix_B}). While training, we perform loss masking for contextual tokens such that only the tokens in the completion part of the sample are learned. This formulation ensures that the model learns to produce a contextually coherent assistant response given the entire preceding dialogue history, without diluting the gradient on earlier turns.

\subsection{Evaluation Protocol}
\label{sec:dyna_eval}

We evaluate a fine-tuned LLM \(f_{\phi}\) along two complementary axes: Static evaluation (isolated response quality) and Dynamic evaluation (end-to-end interactive robustness).

The dialogues produced by the assistant LLM \(f_{\phi}\) are evaluated with four classes of conversation-level metrics: (i) tool-calling and parameter-filling accuracy (\textsc{Acc}); (ii) failure measures (\textsc{FTR}, \textsc{TAR}); (iii) auxiliary metrics: tool-call precision/recall (\textsc{TCP}, \textsc{TCR}) and parameter-key precision/recall (\textsc{PKP}, \textsc{PKR}); and (iv) semantic-fidelity metrics, comprising conversation relevancy (\textsc{ConvRel}), type–token ratio (\textsc{TTR}), and \(n\)-gram diversity (\textsc{NGD}). Verbal definitions of these metrics are given in Section~\ref{sec:experiments}, and their complete mathematical formulations appear in Appendix~\ref{sec:appendix_C1}.


\paragraph{Static Evaluation}
\label{par:stat_eval}

In static evaluation, we sequentially decode each assistant turn \(\hat a_t = f_{\phi}(u_{\le t},\,\hat a_{<t};\,\mathcal{C}_{k})\), leaving user utterances intact. Static evaluation is deterministic, inexpensive, and isolates the model’s ability to emit ``correct’’ replies under perfect user prompts; however, it cannot capture how the assistant’s outputs would influence subsequent user behavior in an interactive setting.

\paragraph{Dynamic Evaluation}
\label{par:dyna_eval}

To gauge \emph{on-policy} conversational competence, the fine-tuned model \(f_{\boldsymbol{\phi}}\) is inserted as the \emph{assistant agent} inside the full \textsc{UTC-Gen} loop (Figure~\ref{fig:data_engine}); the user-proxy policy \(P_{\boldsymbol{\theta}_{u}}\) remains frozen (cf.~§\ref{sec:dialogue_synth}).  The interaction unfolds for at most \(T_{\max}\) turns, yielding a trajectory
\[
  d_{f_{\boldsymbol{\phi}}}
  \;=\;
  \bigl\langle
      (\hat u_{1},\hat a_{1}),
      (\hat u_{2},\hat a_{2}),
      \dots,
      (\hat u_{T'},\hat a_{T'})
  \bigr\rangle,
\]
with \(T' \leq T_{\max}\). At turn \(t\), the assistant observes the dialogue prefix \(\hat{\mathbf{h}}^{a}_{t} = (\hat u_{1},\hat a_{1},\dots,\hat u_{t})\) together with the candidate-tool set \(\mathcal{C}_{k}\) and generates \(\hat a_{t} = f_{\boldsymbol{\phi}}(\hat{\mathbf{h}}^{a}_{t};\,\mathcal{C}_{k})\). This rollout measures the model's ability to maintain contextual coherence, self-correct earlier reasoning errors, and issue schema-conformant tool calls.

\begin{table*}[t]
  \centering
  \small

  \setlength{\tabcolsep}{4pt}
  \begin{tabular}{
      l  |                    
      S 
      S 
      S | 
      S 
      S 
      S 
  }
  \toprule
  \multirow{2}{*}{\textbf{Model}} &
    \multicolumn{3}{c}{\textbf{Static Evaluation}} &
    \multicolumn{3}{c}{\textbf{Dynamic Evaluation}} \\
  \cmidrule(lr){2-4}\cmidrule(lr){5-7}
  & {\textsc{Acc} ($\uparrow$)}
  & {\textsc{FTR} ($\downarrow$)}
  & {\textsc{TAR} ($\downarrow$)}
  & {\textsc{Acc} ($\uparrow$)}
  & {\textsc{FTR} ($\downarrow$)}
  & {\textsc{TAR} ($\downarrow$)} \\
  \midrule

Llama-3.2-DiaFORGE-3B \citep{grattafiori2024llama} & $\num{0.52}$ & $\num{0.12}$ & $\num{0.30}$ & $\num{0.80}$ & $\num{0.08}$ & $\num{0.06}$ \\
Llama-3.3-70B & $\num{0.03}$ & $\textbf{\num{0.00}}$ & $\num{0.97}$ & $\num{0.11}$ & $\textbf{\num{0.02}}$ & $\num{0.88}$ \\
Llama-3.3-70B-fc & $\num{0.22}$ & $\num{0.52}$ & $\textbf{\num{0.01}}$ & $\num{0.30}$ & $\num{0.22}$ & $\textbf{\num{0.01}}$ \\
Llama-3.3-DiaFORGE-70B & $\num{0.38}$ & $\num{0.03}$ & $\num{0.58}$ & $\num{0.79}$ & $\num{0.03}$ & $\num{0.15}$ \\
Llama-xLAM-2-70B-fc-r \citep{prabhakar2025apigen} & $\num{0.48}$ & $\num{0.18}$ & $\num{0.13}$ & $\num{0.51}$ & $\num{0.18}$ & $\num{0.05}$ \\
Llama-3.3-Nemotron-Super-49B \citep{bercovich2025llamanemotronefficientreasoningmodels} & $\num{0.60}$ & $\num{0.07}$ & $\num{0.25}$ & $\num{0.72}$ & $\num{0.08}$ & $\num{0.08}$ \\
Llama-3.3-Nemotron-DiaFORGE-49B & $\textbf{\num{0.82}}$ & $\num{0.04}$ & $\num{0.12}$ & $\textbf{\num{0.89}}$ & $\num{0.06}$ & $\num{0.03}$ \\
\hline
Gemma-3-4B \citep{team2025gemma} & $\num{0.19}$ & $\num{0.17}$ & $\num{0.61}$ & $\num{0.24}$ & $\num{0.14}$ & $\num{0.58}$ \\
Gemma-3-DiaFORGE-4B & $\num{0.53}$ & $\num{0.05}$ & $\num{0.37}$ & $\num{0.81}$ & $\num{0.09}$ & $\num{0.05}$ \\
Gemma-3-12B & $\num{0.31}$ & $\num{0.03}$ & $\num{0.62}$ & $\num{0.37}$ & $\num{0.04}$ & $\num{0.57}$ \\
Gemma-3-DiaFORGE-12B & $\num{0.68}$ & $\textbf{\num{0.02}}$ & $\num{0.26}$ & $\num{0.86}$ & $\num{0.07}$ & $\num{0.07}$ \\
Gemma-3-27B & $\num{0.19}$ & $\num{0.02}$ & $\num{0.78}$ & $\num{0.21}$ & $\textbf{\num{0.00}}$ & $\num{0.79}$ \\
Gemma-3-DiaFORGE-27B & $\textbf{\num{0.77}}$ & $\num{0.03}$ & $\textbf{\num{0.18}}$ & $\textbf{\num{0.89}}$ & $\num{0.03}$ & $\textbf{\num{0.03}}$ \\
\hline
GPT-4o-20241120 \citep{hurst2024gpt} & $\num{0.19}$ & $\num{0.00}$ & $\num{0.81}$ & $\num{0.62}$ & $\num{0.02}$ & $\num{0.36}$ \\
GPT-4o-20241120-fc & $\num{0.61}$ & $\num{0.64}$ & $\num{0.16}$ & $\num{0.56}$ & $\num{0.59}$ & $\num{0.05}$ \\
Claude-3.5-Sonnet-20241022 \citep{claude35_sonnet_2024} & $\num{0.15}$ & $\num{0.02}$ & $\num{0.82}$ & $\num{0.39}$ & $\num{0.03}$ & $\num{0.55}$ \\
Claude-3.5-Sonnet-20241022-fc & $\num{0.42}$ & $\num{0.76}$ & $\num{0.03}$ & $\num{0.40}$ & $\num{0.34}$ & $\num{0.03}$ \\
  \bottomrule
  \end{tabular}
  \caption{\label{tab:tool_eval}Evaluation Results on Tool Call Accuracy and Failure Modes. \textbf{All open-source models evaluated are instruction-tuned, decoder-only LLMs.} Models with the suffix “\textit{fc}” support native function/tool calling, while all other models are evaluated using CAPO-optimized system prompts. Bootstrap 95\,\% CIs for static and dynamic \textsc{Acc} are reported in Appendix~\ref{sec:appendix_C2}.}
  \setlength{\tabcolsep}{6pt}
\end{table*}

\section{Experiments}
\label{sec:experiments}

We fine-tune six publicly available, instruction-tuned, decoder-only language models: Llama-3.2-3B, Gemma-3-4B, Gemma-3-12B, Gemma-3-27B, Llama-3.3-Nemotron-Super-49B, Llama-3.3-70B.

\paragraph{Training Configuration}
All models are fine-tuned exclusively on the 5,000 DiaFORGE conversations, yielding 13,649 turn-sliced completion samples generated by the data engine illustrated in Figure~\ref{fig:data_engine}. No additional general-domain SFT data is incorporated. Each base model is trained for a single epoch using the AdamW optimizer~\cite{loshchilov2017decoupled}. Complete hyperparameter settings and an annotated training sample are provided in Appendix~\ref{sec:appendix_B}.

\paragraph{Evaluation Setting} We evaluate and compare the performance of our fine-tuned models against several baselines: non-fine-tuned models, closed-source models such as \textit{GPT-4o} and \textit{Claude-3.5-Sonnet}, and \textit{Llama-xLAM-2-70b-fc-r}, the state of the art for function calling at the time of writing this paper according to \textsc{BFCL v3}~\citep{yan2024bfcl}. For non-fine-tuned and closed-source models, we apply system prompt optimization using Cost-Aware Prompt Optimization (CAPO)~\citep{zehle2025capo}, the state-of-the-art prompt optimization method at the time of writing (Appendix \ref{sec:appendix_E}).

Our evaluation benchmark, \textsc{DiaBENCH}, comprises 119 seed tools, each paired with corresponding multi-turn, reasoning-annotated dialogues. The benchmark is built from a \emph{proprietary}, out-of-domain corpus tied to a production assistant and includes held-out, out-of-distribution line-of-business (LoB) tools spanning backend APIs and UI-triggered operations that never appear in training data. Section~\ref{sec:comp_diabench_and_bfcl} details the statistical analysis and its difference with the training data for reproducibility purposes. Experiments employ both the \textbf{static} and \textbf{dynamic} protocols defined in §\ref{sec:dyna_eval}.

\paragraph{Evaluation Metrics.}
\label{par:eval_metrics}
We track dialogue‐level measures for each simulated conversation. \emph{Accuracy Rate} (\textsc{Acc}) is the proportion of multi-turn dialogues in which the assistant's first tool invocation (i) correctly selects the reference tool $\tau^\star$ and (ii) supplies the complete, yet no superfluous, set of required key–value arguments. \emph{False-Positive Tool-call Rate} (\textsc{FTR}) captures any instance where the assistant takes an unwarranted action such as invoking a distractor tool, hallucinating a non-existent endpoint, or issuing multiple tool calls when only one is appropriate.  \emph{Tool-call Abstention Rate} (\textsc{TAR}) captures the converse failure mode: cases where a dialogue concludes without any tool invocation, signaling that the model failed to recognize when tool use was necessary. Together, \textsc{FTR} and \textsc{TAR} directly quantify failures in tool disambiguation, a core aspect of our evaluation (see Appendix~\ref{sec:appendix_C}).  To assess dialogue-level quality beyond tool usage, we further report three complementary metrics: \emph{conversation relevancy} (\textsc{ConvRel}), \emph{type–token ratio} (\textsc{TRR}), and \emph{n-gram diversity} (\textsc{NGD}). Formal definitions for all metrics appear in Appendix~\ref{sec:appendix_C1}.

\paragraph{User Agent in Dynamic Evaluation}
In dynamic evaluation (§\ref{sec:dyna_eval}), the LLM acting as the user-proxy agent is susceptible to hallucinations~\cite{huang2025survey}, which can cause cascading failures in dialogue generation. Such conversations are unsuitable for assessing the assistant model, as failures may stem from user-side hallucinations rather than assistant shortcomings. To mitigate this, we adopt a multi-sampling and voting strategy to generate each user utterance, enhancing stability and reducing evaluation noise.  To generate each user utterance, we sample \textbf{3} candidate responses from the same LLM. A separate voting LLM then selects the best response among them. For the evaluations reported in Table~\ref{tab:tool_eval}, we use differently prompted instances of GPT-4o for sampling and voting. A comparative analysis of alternative sampling models is provided in Appendix~\ref{sec:appendix_C3}. Finally, all conversations generated during dynamic evaluation are manually reviewed by domain experts to detect hallucinations introduced by the user-proxy agent. We observe a user-proxy hallucination rate below \textbf{1\%} across all samples; these instances are excluded prior to computing the final evaluation results.

\begin{figure}[t]
  \centering
  \includegraphics[width=\columnwidth]{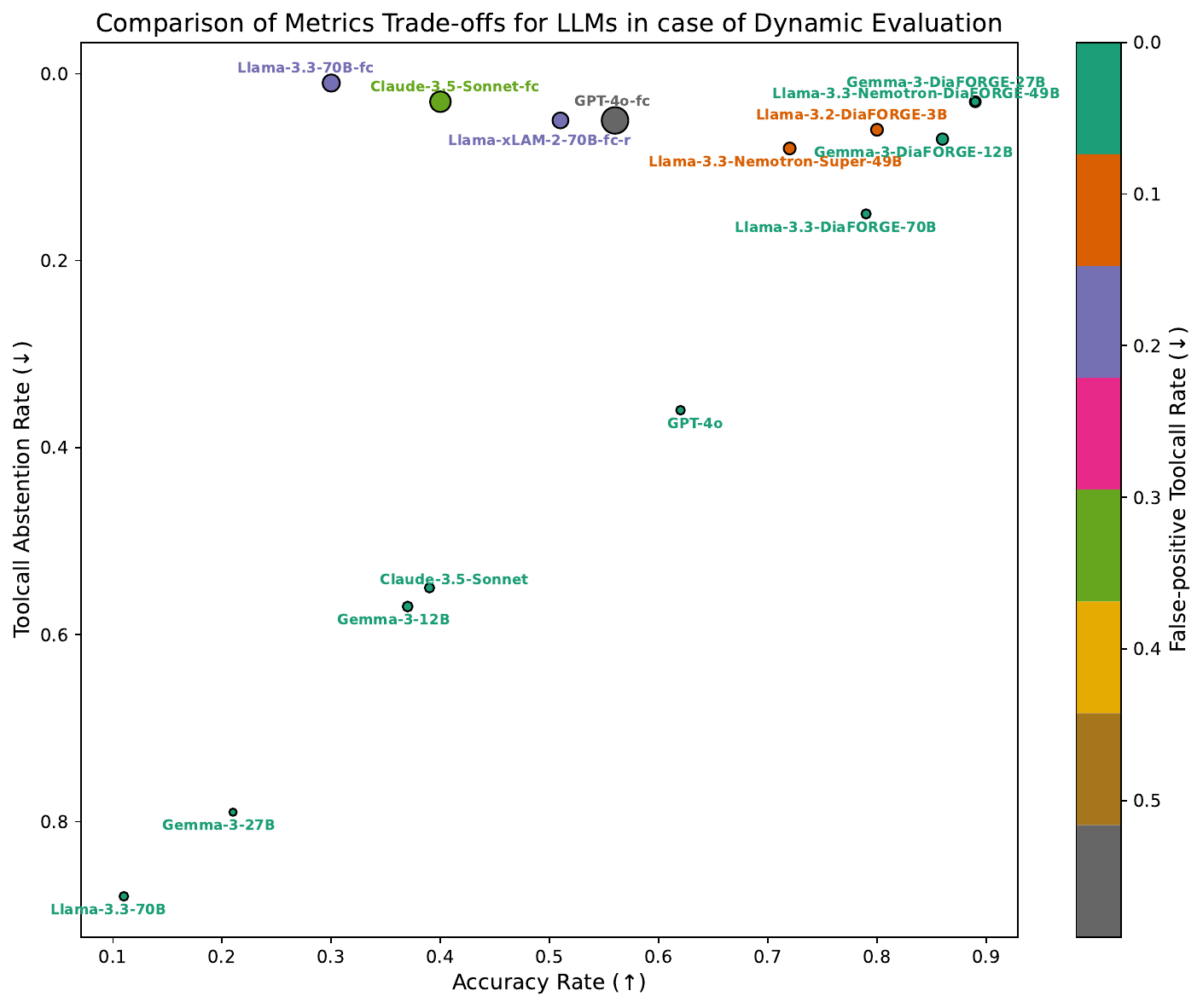}
  \caption{Trade-offs among tool call-related metrics under \textit{Dynamic Evaluation}. 
  Marker size \& Color $\propto$ \textsc{False-Positive Tool-call Rate} (\textsc{FTR}). Models closer to the upper right are preferable; those in the lower left underperform across metrics.}
  \label{fig:pareto_dyneval}
\end{figure}

\begin{table*}[t]
  \centering
  \small
  \setlength{\tabcolsep}{7pt}
  \begin{tabular}{l|ccc|ccc}
    \toprule
    \multirow{2}{*}{\textbf{Setting}} &
    \multicolumn{3}{c}{\textbf{Static Evaluation}} &
    \multicolumn{3}{c}{\textbf{Dynamic Evaluation}} \\
    \cmidrule(lr){2-4}\cmidrule(lr){5-7}
    & {\textsc{Acc} ($\uparrow$)}
    & {\textsc{FTR} ($\downarrow$)}
    & {\textsc{TAR} ($\downarrow$)}
    & {\textsc{Acc} ($\uparrow$)}
    & {\textsc{FTR} ($\downarrow$)}
    & {\textsc{TAR} ($\downarrow$)} \\
    \midrule
    Gemma-3-27B & $\num{0.19}$ & $\textbf{\num{0.02}}$ & $\num{0.78}$ & $\num{0.21}$ & $\textbf{\num{0.00}}$ & $\num{0.79}$ \\
    Gemma-3-DiaFORGE-27B & $\textbf{\num{0.77}}$ & $\num{0.03}$ & $\textbf{\num{0.18}}$ & $\textbf{\num{0.89}}$ & $\num{0.03}$ & $\textbf{\num{0.03}}$ \\
    w/o Validation Cascade & $\num{0.46}$ & $\num{0.04}$ & $\num{0.47}$ & $\num{0.56}$ & $\num{0.06}$ & $\num{0.35}$ \\
    w/o Near-Duplicate Distractor Sampling & $\num{0.52}$ & $\num{0.10}$ & $\num{0.34}$ & $\num{0.63}$ & $\num{0.18}$ & $\num{0.19}$ \\
    w/o Thinking Traces & $\num{0.62}$ & $\num{0.04}$ & $\num{0.30}$ & $\num{0.77}$ & $\num{0.16}$ & $\num{0.04}$ \\
    \bottomrule
  \end{tabular}
  \caption{Ablation study on \textbf{Gemma-3-DiaFORGE-27B}: each variant removes one UTC-GEN component.}
  \label{tab:ablation_g27b}
\end{table*}

In Table~\ref{tab:tool_eval}, we compare all evaluated models using three tool call-related metrics: \textsc{Acc}, \textsc{FTR}, and \textsc{TAR}. These metrics collectively assess an LLM’s ability to invoke tools reliably in realistic, multi-turn settings. \textsc{Acc} measures correctness, \textsc{FTR} captures incorrect tool calls, and \textsc{TAR} reflects the risk of failing to complete the tool-calling objective within the dialogue. For an LLM to be viable in an industry setting, mitigating the risks of insufficient disambiguation, it must balance the three metrics while demonstrating reliability on each. Figure~\ref{fig:pareto_dyneval} illustrates the trade-offs among these metrics for different models. We observe that models trained with \textsc{DiaFORGE} achieve high \textsc{Acc} while simultaneously minimizing both \textsc{FTR} and \textsc{TAR}. Detailed evaluation results for more granular metrics (\textsc{TCP}, \textsc{TCR}, \textsc{PKP}, \textsc{PKR}, etc.) appear in Appendix \ref{sec:appendix_C2}.

At a production scale of 10 k tool-call-eligible conversations per day, even modest differences between LLMs compound into large operational deltas: a GPT-4o-fc configuration yields \textbf{5{,}500--6{,}000} erroneous tool calls per day, resulting in substantial remediation and infrastructure overhead, whereas a GPT-4o (prompt) configuration tends to abstain, stalling \textbf{3{,}500--3{,}800} conversations per day. Both patterns degrade user experience and raise costs, increasing churn. In contrast, \textsc{DiaFORGE}-tuned models reduce total failures to \textbf{250--350} per day, lowering erroneous calls and stalls. Appendix \ref{sec:prod_case_study} provides a real production case study showing impact of such finetuning.


Beyond accurate tool invocation, our use case demands that models also sustain coherent, human-like dialogue throughout the interaction. This includes maintaining context and responding naturally to human users. To assess these capabilities, we report additional metrics related to conversational handling in Appendix~\ref{sec:appendix_C2}.

\paragraph{Disambiguation Frontier.}
\label{par:disambiguation_frontier}
Figure~\ref{fig:frontier} plots dynamic \textsc{Acc} against the cosine similarity of the closest distractor tool across five similarity bins (computed once per test case using \texttt{text-embedding-3-large} embeddings; see Appendix~\ref{sec:appendix_disamb_frontier} for details). \textsc{DiaFORGE}-tuned models (\textit{Gemma-3-DiaFORGE-27B}, \textit{Nemotron-DiaFORGE-49B}) maintain near-flat accuracy even as distractor similarity approaches 1.0, demonstrating robust disambiguation. By contrast, \textit{GPT-4o} shows marked degradation at high similarity, whereas \textit{Claude-3.5-Sonnet} and \textit{Gemma-3-27B} perform poorly across the entire spectrum. This confirms that \textsc{DiaFORGE} finetuning specifically teaches models to handle fine-grained near-duplicate tool disambiguation rather than relying on surface-level tool name differences.

\begin{figure}[t]
  \centering
  \includegraphics[width=\columnwidth]{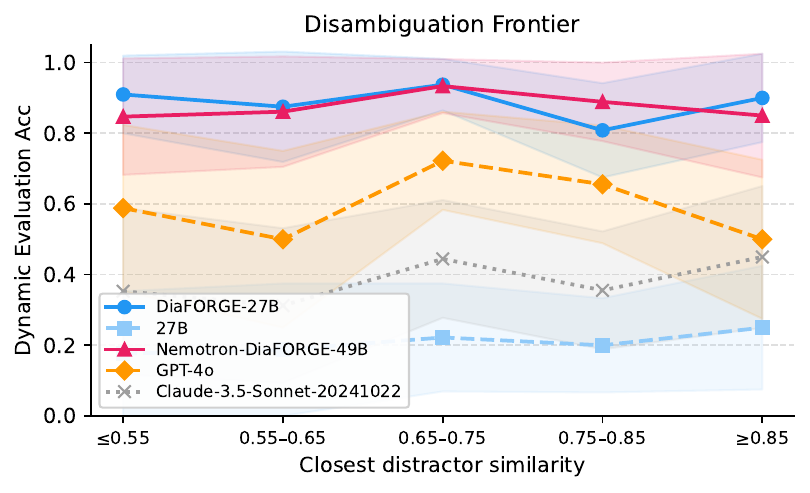}
  \caption{Disambiguation frontier: dynamic evaluation accuracy vs.\ closest distractor similarity (5 similarity bins). Shaded bands are bootstrap 95\,\% CIs.}
  \label{fig:frontier}
\end{figure}



\subsection{Ablation Study}
\label{sec:ablation_g27b}

\noindent We run ablations on \textbf{Gemma-3-27B}, holding constant data volume, LoRA recipe, and evaluation protocol; each variant removes exactly one \textsc{UTC-GEN} component to isolate its contribution to disambiguation in the fine-tuned model:
\begin{itemize}
  \item \textbf{Without Validation Cascade:} removes the rule-based and LLM validators; synthetic dialogues enter the training corpus unvalidated.
  \item \textbf{Without Near-Duplicate Distractor Sampling:} removes near-duplicate distractor tools from retrieved tool sets; eliminates turns devoted to fine-grained tool disambiguation.
  \item \textbf{Without Thinking Traces:} removes the assistant’s reasoning traces during fine-tuning and decode without thinking at inference.
\end{itemize}

\noindent Table~\ref{tab:ablation_g27b} compares \emph{Gemma-3-DiaFORGE-27B} with the vanilla backbone and three ablations. The \textsc{DiaFORGE} model attains high \textsc{Acc}, reducing abstention and keeping erroneous calls low. Dropping the \emph{validation cascade} admits schema-invalid or tool-absent turns into training, regressing performance and inflating \textsc{TAR}. Removing \emph{near-duplicate distractor sampling} weakens supervision for fine-grained tool selection and degrades performance on \textsc{DiaBENCH}, which explicitly stresses near-duplicate disambiguation. Ablating \emph{thinking traces} reduces \textsc{Acc}, increasing both erroneous invocations and unnecessary abstentions. Collectively, these results show that validation filters keep training data clean, near-duplicate sampling teaches disambiguation, and reasoning traces calibrate inference toward correct actions.

\subsection{Comparing \textsc{DiaBENCH} with Public Benchmarks}
\label{sec:comp_diabench_and_bfcl}

\paragraph{BFCL v3} BFCL ~\citep{yan2024bfcl} is a general function-calling benchmark that evaluates both native and prompt-induced tool calling. It spans five high-level categories with 17 subcategories; only \textbf{7} subcategories present multiple tools in the retrieval set (LLM context). Across the multi-tool test cases (\textbf{41.2\%} of all samples), the average number of distractor tools is \(\sim\)\textbf{1.6} with an average semantic overlap of \textbf{0.24}. Only \textbf{0.57\%} of all test cases contain at least one distractor that is a \emph{near-duplicate} of the ground-truth tool (similarity criterion in Appendix~\ref{sec:appendix_A4}). Furthermore, \textsc{BFCL v3} adopts a \emph{static} evaluation protocol that follows fixed conversation scripts and assumes fully specified queries, leaving no scope for the model to ask clarifying questions.

Other function-calling benchmarks like ToolBench \cite{qin2023toolllm} \& API-Bank \cite{li2023apibank} also follow similar pattern (static scripts with fully specified queries) and does not evaluate disambiguation capability of the model. To the best of our knowledge, there is no open benchmark yet that evaluates LLM's ability to ask clarifying questions. Thus, we do not expect our finetuned model to outperform on these public benchmarks. However, in order to verify that the finetuning doesn't lead to catastrophic forgetting, we perform parity checking of the \textsc{DiaFORGE}-finetuned model w.r.t instruction-tuned base model on \textsc{BFCL v3}~\citep{yan2024bfcl} and \textsc{MT-Bench}~\citep{zheng2023judging}. The evaluation results (Appendix \ref{sec:bfcl}) suggests no significant degradation in model's general abilities when finetuned with \textsc{DiaFORGE} pipeline.

\paragraph{\textsc{DiaBENCH}} By contrast, \textsc{DiaBENCH} is expressly designed to test an LLM’s ability to pose clarifying questions and disambiguate among near-duplicate tools in \emph{enterprise setting}. \textbf{100\%} of test cases include multiple tool options in context, with an average of \(\sim\)\textbf{5.2} distractor tools per case and an average retrieval-set semantic overlap of \textbf{0.47}. Additionally, \textbf{29.2\%} of all test cases contain distractors that are near-duplicates of the ground-truth tool (same criterion as Appendix~\ref{sec:appendix_A4}), and \textbf{75.6\%} of test cases have parameterized ground truth tools requiring the assistant LLM to ask a clarifying question to obtain missing required parameters. This distribution is intentional and mirrors production traffic. Beyond static scoring, we also employ a \emph{dynamic} protocol that redeploys each post-trained model in an agentic loop to assess whether it proactively solicits missing information via clarifying questions and subsequently executes the correct tool call.

\begin{figure}[ht]
\centering
\includegraphics[width=0.7\columnwidth, trim = 0.5cm 0.5cm 0.5cm 0.5cm, clip]{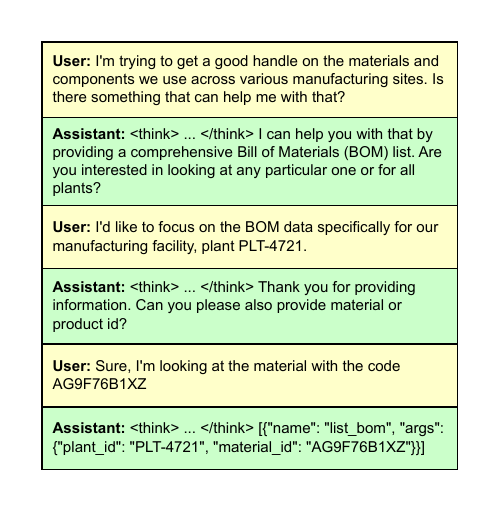}
\caption{Example \textsc{DiaFORGE} training dialogue generated with the UTC-GEN user proxy.}
\label{fig:diaforge_train_dialogue}
\end{figure}

\begin{figure}[ht]
\centering
\includegraphics[width=0.7\columnwidth, trim = 0.5cm 0.5cm 0.5cm 0.5cm, clip]{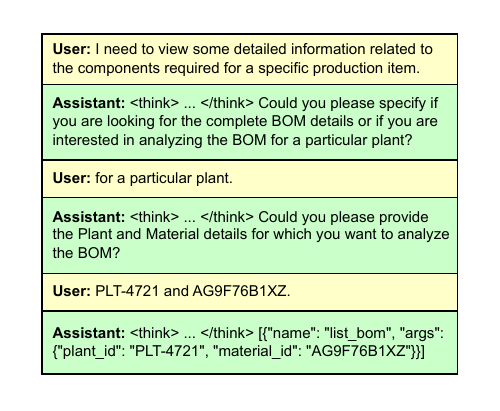}
\caption{Example dialogue from dynamic evaluation: the user proxy LLM is prompted to respond concisely; the assistant LLM is the fine-tuned \textsc{Gemma-3-DiaFORGE-27B}.}
\label{fig:diabench_samples}
\end{figure}

Unlike \textsc{BFCL v3}, \textsc{DiaBENCH} comprises complex business functions spanning backend APIs from \emph{held-out} lines of business (LoBs), high-level workflow functions, and UI-triggering functions. Our production environment covers nine LoBs with thousands of backend APIs. For training, we stratified-sample backend APIs from five LoBs. For evaluation, \textsc{DiaBENCH} is constructed from LoB functions \emph{disjoint} from those used in training and includes workflow/UI functions that \textbf{never} appear in the training corpus.

User-proxy prompting intentionally differs between the training corpus and \textsc{DiaBENCH}. As shown in Figure~\ref{fig:diaforge_train_dialogue}, the training user proxy (via \textsc{UTC-GEN}) produces verbose, often ambiguous utterances, whereas in \textsc{DiaBENCH} dynamic evaluation (Figure~\ref{fig:diabench_samples}) the dedicated user proxy is prompted to issue concise, command-like queries. This deliberate style shift induces out-of-distribution, terse inputs that stress an LLM's ability to ask targeted clarifying questions, ultimately select the correct tool, and closely reflects the interaction patterns seen in real user behavior with our AI assistant in production.
We provide detailed construction guidelines in Appendix~\ref{sec:diabench_guidelines}, including how to build a comparable benchmark directly from our released open corpus.

\section{Conclusion}

We introduce \textsc{DiaFORGE}, a modular three-stage pipeline that (i) synthesizes high-quality, multi-turn tool-calling dialogues designed to stress the disambiguation behaviors where current LLMs still struggle, (ii) enables efficient supervised fine-tuning across models of varying scales, and (iii) provides both static and dynamic evaluation tailored to realistic multi-turn tool use in enterprise settings. To spur further research on robust, real-world tool-calling agents, we publicly release a dataset of roughly \(5{,}000\) production-grade enterprise APIs paired with their \textsc{DiaFORGE}\nobreakdash-curated dialogues.

\section*{Limitations}
\label{sec:limitations}

DiaFORGE’s \emph{disambiguation-centric} data synthesis paradigm provides a principled foundation for aligning tool invocation with user intent, yet several open challenges remain, which we plan to explore as future work.

Our post-training setup assumes the ground-truth tool is present in the retrieved candidate set: an assumption that does not always hold in production. Future work will incorporate hard negatives and explicit ``no-tool'' dialogues to train the agent to refrain from using tools in such cases.

DiaFORGE uses LLM-based validators to filter unrealistic dialogues, yet these validators can exhibit biases, hallucinate, or miss edge cases. Moreover, the current generators do not yet cover the full breadth of complex enterprise interactions. Strengthening diversity via more robust ensemble validation and expanding generator coverage is a key direction for future work. Meanwhile, extending DiaFORGE to synthesize multi-tool, multi-step, disambiguation-aware conversations would further improve data realism and furnish a more rigorous benchmark of an LLM's ability to plan, sequence, and recover across near-duplicate tools.

The primary reproducibility limitation is \textsc{DiaBENCH}: as a proprietary, production-internal benchmark, the exact evaluation figures cannot be independently replicated without equivalent enterprise data. We mitigate this through detailed construction guidelines (Appendix~\ref{sec:diabench_guidelines}), which show how to build a comparable benchmark from our released open corpus.

Although dynamic evaluation is overall a better strategy to evaluate conversational LLMs, we still require human validation to discard dialogues where the simulated user hallucinates. Such manual validation of the synthesized dialogues during dynamic evaluation is expensive \& hard to scale, especially in an industry setting. Moreover, while our multi-sampling voting strategy tries to minimize the user-proxy hallucination, it leads to an increase in cost due to multiple LLM calls.

\section*{Ethical Considerations}

We conducted experiments within the provisions of the ACL Ethics Policy and relevant research-integrity guidelines. There are, to the best of our knowledge, no remaining ethical risks that have not been addressed.

\bibliography{custom}
\newpage
\appendix

\section{Details About Data Generation Engine}
\label{sec:appendix_A}

\subsection{Dialogue Synthesis}
\label{sec:appendix_A1}

The Multi-Agent Dialogue Synthesizer (Figure~\ref{fig:data_engine}) generates synthetic dialogues in two stages, guided by dialogue state tracking to ensure coherence and goal alignment.

\paragraph{Tool Selection Stage.}
The user-proxy agent is assigned a goal and generates vague but contextually relevant utterances. Its context includes the seed tool $\tau^\star$ and a set of distractor tools $\mathcal{D}_{k}(\tau^\star)$. In this stage, the user-proxy agent is instructed to reveal minimal information initially, offering substantive details only in response to the assistant’s clarifying questions in subsequent turns.

On the other hand, the assistant agent’s objective in this stage is to identify the appropriate tool $\tau^\star$ by asking clarifying questions. It does not have direct access to $\tau^\star$ but instead queries a vector database via a live tool retriever to obtain a candidate set $\mathcal{C}_{k}$. We enforce that $\tau^\star \in \mathcal{C}_{k}$. If the condition is not satisfied, we discard the current conversation and regenerate a new one from scratch with the same seed tool, up to five attempts, to maximize its chances of being included in the training corpus.

Once a tool is selected, a rule-based validation will ensure that the selected tool $\tau = \tau^\star$. If $\tau \neq \tau^\star$, the dialogue sample is rejected and synthesis halts. If $\tau = \tau^\star$, the last assistant message is removed, and the process transitions to the parameter filling stage.

\paragraph{Parameter Filling Stage.}
Assuming the correct tool has been selected, the assistant agent now proceeds to collect the necessary parameters to execute the tool call. With the gold tool $\tau^\star$ provided in its context, the assistant is now tasked with eliciting all required argument values, whether stated explicitly or implied by the user-proxy, and invoking the tool once all required inputs have been gathered.

On the other hand, the user-proxy agent is given access to the ground-truth parameter values, represented as the argument map \(\mathcal{V}^\star = \mathcal{V}\bigl(\tau^{\star}, p\bigr)\), and is instructed to provide the parameter values specifically requested by the assistant. If the selected tool requires no parameters, the assistant initiates the tool call immediately at the beginning of the parameter filling stage, without any additional input from the user-proxy.

Throughout both stages, assistant messages include internal reasoning traces generated using the \textit{Reason First, Response Later} strategy. These traces are accessible only to the assistant and remain hidden from the user-proxy.


\begin{tcolorbox}[enhanced jigsaw,breakable,pad at break*=1mm, colback=yellow!5!white,colframe=yellow!50!black,
  colbacktitle=yellow!75!black,title=Generated Sample]
  \textbf{Persona:} A logistics operations manager seeking advice on mitigating transport risks \\
  \textbf{Seed:} fn\_1126\_cloud\_transport\_management
  \tcblower
  \textbf{User:} I'm trying to improve how we handle different aspects of our transport processes. Could you assist me with finding the right approach or tool to manage and monitor our logistics tasks effectively? \\
  \textbf{Assistant:} <think> ... </think> Could you please specify what aspect of logistics you are trying to manage? Are you focusing on tracking logistics orders, managing warehouse resources, or monitoring transport records? \\
  \textbf{User:} I'm looking for a way to monitor and review all the actions taken during our transport activities. What options are available for keeping an eye on these processes? \\
  \textbf{Assistant:} <think> ... </think> To assist you in monitoring and reviewing all actions taken during your transport activities, our designated tool can be very helpful. Before proceeding, could you please provide the Node ID and the Transport Request ID? These two pieces of information are necessary to retrieve the relevant logs and facilitate monitoring and troubleshooting efficiently. \\
  \textbf{User:} Certainly. The Node ID is 437292, and the Transport Request ID is 957841. Let me know if there's anything else you need. \\
  \textbf{Assistant:} <think> ... </think> [\{"name": "fn\_1126\_cloud\_transport\_management", "args": \{"nodeId": 437292, "transportRequestId": 957841\}\}]
\end{tcolorbox}

\noindent\begin{minipage}{\columnwidth}
\captionof{figure}{\textsc{DiaFORGE} generated dialogue sample}
\label{fig:generation_example}
\end{minipage}

An example of a synthesized dialogue is shown in Figure \ref{fig:generation_example}. The assistant strategically asks specific, targeted questions to progressively narrow down the tool selection. Once the correct tool is identified, it proceeds to elicit the necessary parameter values before issuing a final tool call. In real-world applications, such disambiguation capability is essential for function-calling models to be genuinely helpful and reliable in assisting enterprise users.

\subsection{Dialogue Validation}
\label{sec:appendix_A2}

Once the dialogues are synthesized, they are processed by the Multi-Agent Dialogue Validator, illustrated in Figure~\ref{fig:data_engine}. This system comprises multiple validator agents, broadly categorized into two types.

\paragraph{Functional Validators.} These are rule-based agents designed to enforce structural and logical constraints on the generated dialogue. Multiple functional validators are applied sequentially. \textsc{Format Validator} ensures the dialogue follows the expected structure, alternating user and assistant turns, and that assistant messages include both reasoning traces and final responses. \textsc{Toolcall Validator} verifies that the dialogue ends with a valid tool call corresponding to the gold tool $\tau^\star$. \textsc{Toolargs Validator} checks that all required parameters for the tool call are correctly provided.  Due to interdependencies among these checks, the functional validators are executed in the following order: \textsc{Format Validator} $\rightarrow$ \textsc{Toolcall Validator} $\rightarrow$ \textsc{Toolargs Validator}.

\paragraph{LLM Validators.} These are LLM-based agents responsible for validating aspects that require natural language understanding. Each validator is prompted with distinct instructions and assesses different aspects of the dialogue. \textsc{Relevancy Validator} evaluates whether the dialogue content is semantically relevant to the gold tool $\tau^\star$. \textsc{LLM Critique} assesses the overall flow of the conversation, ensuring it exhibits the expected two-stage structure, and checks that both agents (user and assistant) adhere to their designated roles. As the validators function independently, they are executed concurrently. A dialogue sample is rejected if any validator flags it as invalid, as all validators are considered equally authoritative.

\paragraph{Human Spot Checks.} To complement automated validation, we periodically conduct human spot checks on random subsets of the validated dialogues, providing an additional quality‐control layer and guiding prompt refinements when systematic issues are discovered.

\subsection{Data Distribution}
\label{sec:appendix_A3}

We present the distribution of the training data used in this study, which is identical to the dataset we release as part of our open-sourced benchmark.

Figure~\ref{fig:data_num_turns} illustrates the distribution of conversation lengths, measured by the number of dialogue turns. The majority of conversations contain fewer than five turns, aligning with typical session lengths observed in real-world enterprise tool-use scenarios. Figure~\ref{fig:data_num_params} shows the distribution of the number of parameters associated with the seed tools for which the conversations were generated. 

Figure~\ref{fig:data_num_tool_param_disamb_turns} depicts the number of dialogue turns dedicated to tool disambiguation and parameter filling. In most cases, tool selection is completed within two turns, followed by a single turn for parameter specification. Notably, some samples contain zero turns for parameter filling: this occurs when the tool either requires no parameters or when parameters are provided during the tool selection phase, which reflects common patterns observed in real-world multi-turn enterprise interactions.

\begin{figure}[ht]
\centering
\includegraphics[width=\columnwidth]{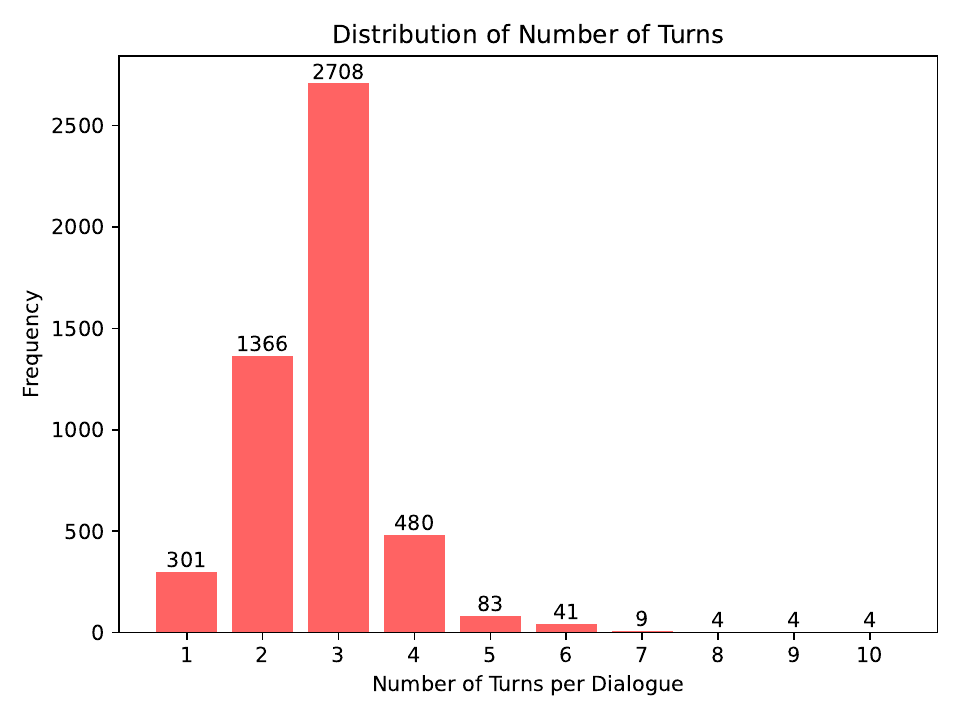}
\caption{Conversation length distribution: number of dialogue turns per sample.}
\label{fig:data_num_turns}
\end{figure}

\begin{figure}[ht]
\centering
\includegraphics[width=\columnwidth]{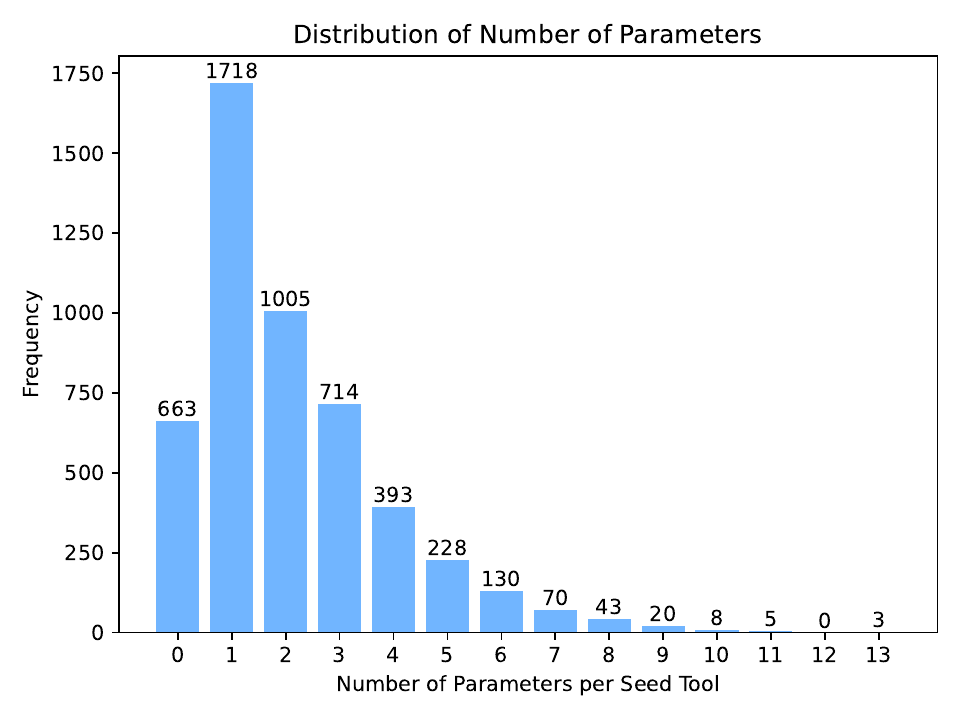}
\caption{Parameter count distribution: number of parameters per seed tool.}
\label{fig:data_num_params}
\end{figure}

\begin{figure*}[ht]
  \includegraphics[width=0.48\linewidth]{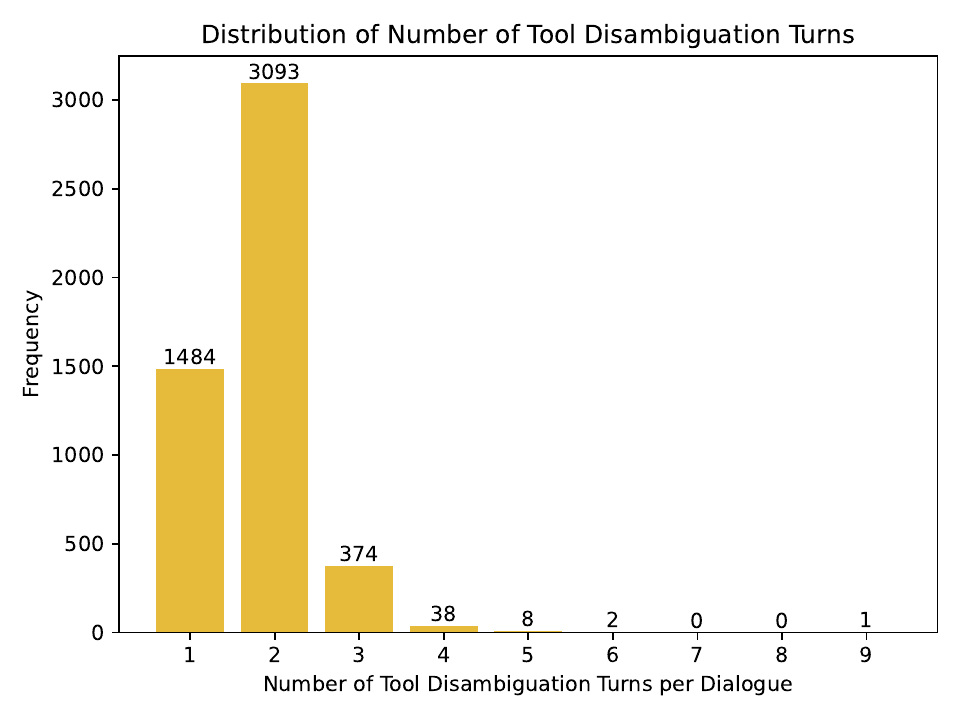} \hfill
  \includegraphics[width=0.48\linewidth]{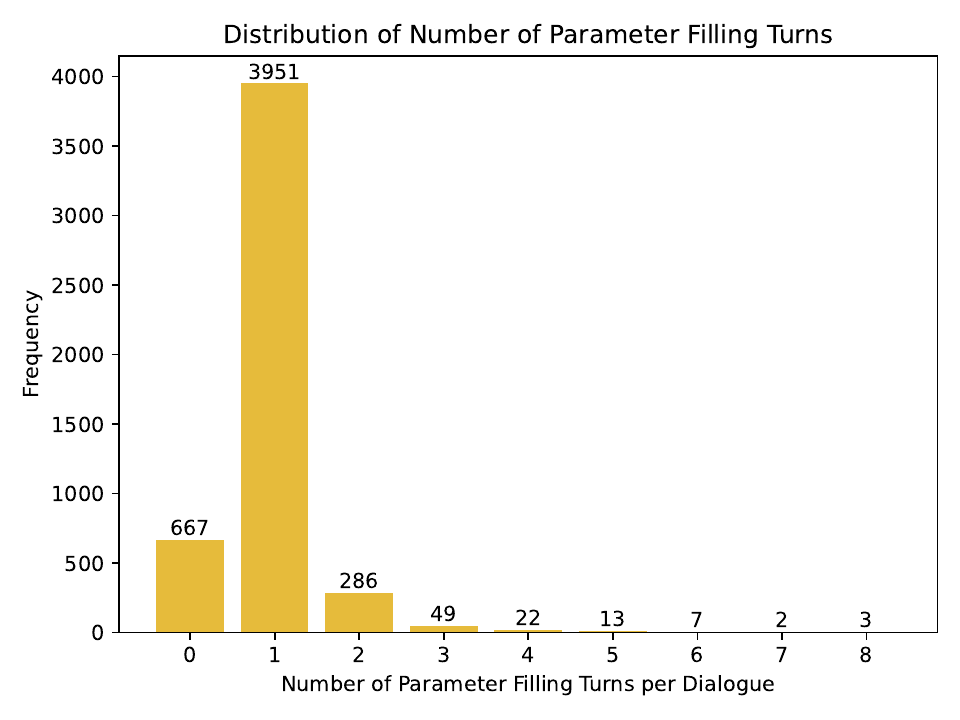}
  \caption{Turn distribution for tool disambiguation (left) and parameter specification (right).}
  \label{fig:data_num_tool_param_disamb_turns}
\end{figure*}

\subsection{Analyzing Near-Duplicate Tools}
\label{sec:appendix_A4}}

We quantify near-duplication with a bounded, symmetric composite similarity metric. Each tool $\tau\in\mathcal{T}$ is represented as $(\texttt{tname}(\tau),\,\texttt{tdesc}(\tau),\,\texttt{params}(\tau))$, where $\texttt{params}(\tau)$ is a JSON Schema map from argument keys to $(\texttt{type},\,\texttt{description},\,\texttt{required})$. Define the set of required keys as $\texttt{keys}(\tau)$.
For any argument $p\in\texttt{keys}(\tau)$, let $\texttt{type}_{\tau}(p)$ denote the normalized base type (e.g., \texttt{string}, \texttt{integer}, \texttt{float}, \texttt{date}, \texttt{bool}). Unless noted, all component similarities lie in $[0,1]$ and are symmetric, and the composite score inherits these properties.

\paragraph{Composite similarity.}
For $\tau\neq\tau^\star$,
\[
\label{eq:composite}
\begin{aligned}
S(\tau^\star,\tau)
&= w_{\mathrm{tname}}\,S_{\mathrm{tname}}(\tau^\star,\tau) \\
&\quad + w_{\mathrm{tdesc}}\,S_{\mathrm{tdesc}}(\tau^\star,\tau) \\
&\quad + w_{\mathrm{param}}\,S_{\mathrm{param}}(\tau^\star,\tau),
\end{aligned}
\]
with $w_{\mathrm{tname}},w_{\mathrm{tdesc}},w_{\mathrm{param}}\in[0,1]$ and $w_{\mathrm{tname}}+w_{\mathrm{tdesc}}+w_{\mathrm{param}}=1$. We use
$w_{\mathrm{tname}}=0.40$, $w_{\mathrm{tdesc}}=0.35$, $w_{\mathrm{param}}=0.25$.

\paragraph{Name similarity.}
Let $\operatorname{LCS}(\cdot,\cdot)$ be the character-level longest common subsequence. With preprocessed names (lowercased), define
\[
\label{eq:sname}
S_{\mathrm{tname}}(\tau^\star,\tau)
~=~
\frac{2\,\operatorname{LCS}\!\bigl(\texttt{tname}(\tau^\star),\texttt{tname}(\tau)\bigr)}
     {\lvert \texttt{tname}(\tau^\star)\rvert + \lvert \texttt{tname}(\tau)\rvert}.
\]

\paragraph{Description similarity.}
Let $\psi(\cdot)$ be a sentence encoder and define unit vectors
$\mathbf{v}(\tau)=\psi\!\bigl(\texttt{tdesc}(\tau)\bigr)/\bigl\|\psi\!\bigl(\texttt{tdesc}(\tau)\bigr)\bigr\|_2$
so that $\|\mathbf{v}(\tau)\|_2=1$. Cosine lies in $[-1,1]$; we rescale to $[0,1]$:
\[
\label{eq:sdesc}
S_{\mathrm{tdesc}}(\tau^\star,\tau)
~=~
\frac{1+\mathbf{v}(\tau^\star)^\top \mathbf{v}(\tau)}{2}.
\]

\paragraph{Parameter similarity.}
Write $A=\texttt{keys}(\tau^\star)$, $B=\texttt{keys}(\tau)$, and $I=A\cap B$. Combine set overlap with type agreement:

\[
\label{eq:sset}
S_{\mathrm{set}}(\tau^\star,\tau)=
\begin{cases}
\dfrac{|A\cap B|}{|A\cup B|}, & \text{if } A\cup B\neq\varnothing,\\[4pt]
1, & \text{if } A=B=\varnothing.
\end{cases}
\]

\[
\label{eq:stype}
S_{\mathrm{type}}(\tau^\star,\tau)
= \frac{1}{|I|}
\sum_{p\in I}
\mathbb{I}\!\bigl[\texttt{type}_{\tau^\star}(p)=\texttt{type}_{\tau}(p)\bigr].
\]
\noindent\emph{Convention:} if $I=\varnothing$, the sum is $0$ and the ratio above is defined to be $0$ (empty-average).

\[
\label{eq:sparam}
S_{\mathrm{param}}(\tau^\star,\tau)
= \tfrac{1}{2}\,S_{\mathrm{set}}(\tau^\star,\tau)
+ \tfrac{1}{2}\,S_{\mathrm{type}}(\tau^\star,\tau).
\]

\paragraph{Decision rule.}
Flag $\tau$ as a near duplicate of $\tau^\star$ iff
\[
\label{eq:threshold}
S(\tau^\star,\tau)\;\ge\;t,
\qquad t=0.70.
\]

\subsection{DiaBENCH Construction Guidelines}
\label{sec:diabench_guidelines}

A comparable benchmark can be built directly from our openly released corpus of \(\sim\!5{,}000\) enterprise APIs and \textsc{UTC-GEN}-curated dialogues: simply hold out a disjoint subset of tools as seed tools and follow the steps below.

\begin{enumerate}[label=(\arabic*),leftmargin=*,itemsep=2pt]
  \item \textbf{Tool selection.} Sample seed tools exclusively from a held-out split of the API catalogue (disjoint from those used during fine-tuning), ensuring coverage of varied function types.
  \item \textbf{Distractor assembly.} For each seed tool, retrieve $k{=}5$ semantic nearest neighbors using the composite similarity metric (\S\ref{sec:appendix_A4}); target $\geq\!25\%$ of test cases to contain at least one near-duplicate ($S\!\geq\!0.70$) to stress fine-grained disambiguation.
  \item \textbf{Dialogue generation.} Synthesize multi-turn dialogues via \textsc{UTC-GEN} with a \emph{concise, command-like} user-proxy prompt, distinct from the verbose training queries, to induce out-of-distribution generalization.
  \item \textbf{Parameter coverage.} Ensure $\geq\!70\%$ of test cases require at least one clarifying question to elicit missing required parameters, mirroring realistic production traffic.
  \item \textbf{Validation.} Apply the Validator Cascade (\S\ref{par:validator}) and supplement with expert manual review.
\end{enumerate}

\section{Details About Supervised Fine-Tuning}
\label{sec:appendix_B}

We perform Supervised Fine-Tuning (SFT) on top of open-source models that have already been instruction-tuned. While such models are generally optimized across a range of instruction-following tasks, our objective is to further specialize them for tool-calling use cases, enhancing both reliability and usability in enterprise scenarios.

\begin{figure}[ht]
\centering
\includegraphics[width=\columnwidth, trim = 0cm 0.5cm 0cm 0.5cm, clip]{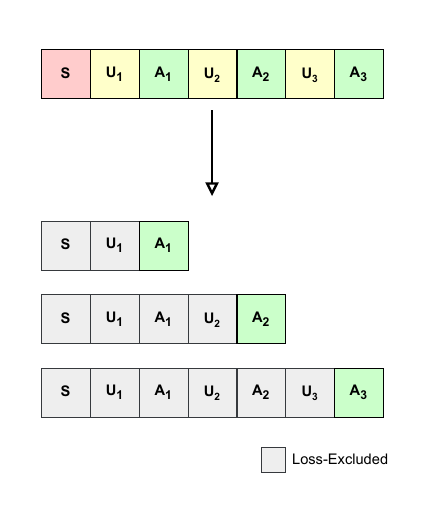}
\caption{Turn slicing and loss masking strategy for SFT sample preparation}
\label{fig:data_prep_sft}
\end{figure}

Figure~\ref{fig:data_prep_sft} illustrates the data preparation pipeline for SFT. We apply a turn-slicing strategy to the synthetic multi-turn dialogues generated by our data engine: for a dialogue consisting of $L_t$ turns, we create $L_t$ separate training samples, each corresponding to an individual assistant response. This allows the model to learn assistant behavior in a fine-grained, turn-wise manner.

For each of these training samples, we apply loss masking, such that only the final assistant message in the sliced context contributes to the training loss. This prevents the model from overfitting to preceding system or user messages and instead focuses learning on assistant behavior. \citep{shi2024instruction} showed that eliminating loss masking, thereby fine-tuning on system \& user instructions, benefits single-turn dialogue tasks, but our empirical observations shows that applying this tactic to multi-turn settings has the opposite effect: the overwhelming volume of unmasked system \& user tokens skews the training signal and noticeably degrades assistant performance at inference.


We fine-tune the models using \textbf{Lo}w-\textbf{R}ank \textbf{A}daptation (LoRA) with a rank of $r = 16$ and a scaling factor $\mathit{\alpha} = 16$. Training is conducted for a single epoch using 8-bit precision and a completion batch size of 1, where each batch consists of one assistant response (as the output) along with its associated metadata and dialogue history (as input). We employ the \textsc{AdamW} optimizer with a peak learning rate of $10^{-4}$ and a cosine learning rate schedule.

\section{In-Depth Analysis of Evaluation}
\label{sec:appendix_C}

This appendix decomposes the composite tool-calling metrics introduced in \S\ref{par:eval_metrics} into atomic, interpretable measures and reports the corresponding results.  We also introduce auxiliary conversational metrics that probe other aspects of agent behavior.

\newcommand{\ind}[1]{\mathbf{1}\!\bigl[#1\bigr]}
\subsection{Evaluation Metrics}
\label{sec:appendix_C1}

Let the evaluation set be
\(
  \mathcal{S}=\{d^{(1)},\dots,d^{(N)}\}.
\)
For a dialogue
\(
  d = \langle (u_{1},a_{1}),\dots,(u_{T},a_{T})\rangle,\ 1\!\le\!T\!\le\!T_{\max},
\)
denote the \emph{reference} tool specification by
\[
  g(d)=\bigl(\{\tau^{\star}(d)\},\;\{\tau^{\star}(d)\;\longrightarrow\;\mathcal{V}^{\star}(d)\}\bigr),
\]
where \(\tau^{\star}(d)\in\mathcal{T}\) is the unique gold tool and \(\mathcal{V}^{\star}(d): (\textsc{Key}\!\to\!\textsc{Value})\) is the corresponding ground-truth map of required arguments.

For any assistant utterance \(a_{t}\) we define
\begin{align*}
  &\mathrm{tools}(a_{t}) \subseteq \mathcal{T}, \\
  \mathrm{args}(a_{t})\;:\;&\mathrm{tools}(a_{t})\;\longrightarrow\;(\textsc{Key}\!\to\!\textsc{Value}),
\end{align*}
where \(\mathrm{tools}(a_{t})\) is the set of tool-identifiers invoked at turn \(t\), and \(\mathrm{args}(a_{t})\) provides a corresponding argument map for each tool in this set. Whenever \(\mathrm{tools}(a_{t})=\varnothing\), \(\mathrm{args}(a_{t})=\varnothing\).

Define the first tool-bearing turn
\[
  t^{\dagger}
  ~=~
  \min\,\bigl\{\,t \mid \mathrm{tools}(a_{t})\neq\varnothing\bigr\},
\]
with the convention \(t^{\dagger}=+\infty\) if the dialogue contains no tool call.  For a dialogue $d$, let
\[
  c(d)=
  \begin{cases}
    \bigl(\mathrm{tools}(a_{t^{\dagger}}),\;
          \mathrm{args}(a_{t^{\dagger}})\bigr),
    & t^{\dagger}<\infty,\\[6pt]
    \varnothing,
    & t^{\dagger}=\infty.
  \end{cases}
\]
Corpus-level prediction and reference tool calls can subsequently be aggregated into:
\[
  C \;=\; \bigl\{\,c(d) \,\bigm|\, d\in\mathcal{S}\bigr\},
  \quad
  G \;=\; \bigl\{\,g(d) \,\bigm|\, d\in\mathcal{S}\bigr\}.
\]
We then construct an alignment multiset that pairs each prediction with its corresponding reference:
\begin{align*}
  \mathcal{M}
  \;&=\;
  \bigl\{
    \bigl(c(d),g(d)\bigr)
    \ \bigm|\ 
    d\in\mathcal{S},\ \\
    \!c(d)&\neq\varnothing,\ 
    \tau^{\star}(d) \in \texttt{tnames}\!\bigl(c(d)\bigr)
  \bigr\}.
\end{align*}
Here, \(\texttt{tnames}(\,\cdot\,)\) returns the set containing the invoked tool-identifiers.  Each predicted call is matched to the unique reference call from the same dialogue \emph{iff} both invoke the identical set of tool-identifiers; otherwise the prediction remains unaligned.  Analogously, \(\texttt{keys}(\,\cdot\,)\) returns the set of argument-key names supplied in the call.

\paragraph{Dialogue–Level Indicators.} For every conversation \(d\in\mathcal{S}\), we compute three indicators:
\begin{itemize}[leftmargin=1.7em,itemsep=3pt]
  \item \textbf{Tool-Call Accuracy} (\textsc{Acc}).  
        The model’s invocation matches the reference tool \emph{and} its full key–value argument map:
        \[
          \textsc{Acc}(d)
          \;=\;
          \ind{c(d)=g(d)}.
        \]
  \item \textbf{False-Positive Tool-Call} (\textsc{FTR}). 
        A tool call is made, but the invoked tool-identifier deviates from the reference:
        \[
        \textsc{FTR}(d)=
        \begin{cases}
        \displaystyle
        \sum_{\tau\in\mathrm{tools}(a_{t^{\dagger}})}
          \ind{\tau \neq \tau^{\star}}, &t^{\dagger}<\infty,\\[9pt]
        0, &t^{\dagger}=\infty.
        \end{cases}
        \]
        If the assistant predicts more than one tool, every superfluous invocation is counted toward the \textsc{FTR} metric.
  \item \textbf{Tool-Call Abstention} (\textsc{TAR}).  
        The dialogue terminates without any tool invocation:
        \[
          \textsc{TAR}(d)
          \;=\;
          \ind{c(d)=\varnothing}.
        \]
\end{itemize}

\paragraph{Corpus-Level Aggregation.} Let
\begin{align*}
  \textsc{Acc} = \frac{1}{|\mathcal{S}|}\sum_{d\in\mathcal{S}}\textsc{Acc}(d), \\
  \textsc{FTR} = \frac{1}{|\mathcal{S}|}\sum_{d\in\mathcal{S}}\textsc{FTR}(d), \\
  \textsc{TAR} = \frac{1}{|\mathcal{S}|}\sum_{d\in\mathcal{S}}\textsc{TAR}(d).
\end{align*}
Together, \textsc{Acc} gauges correct disambiguation and slot filling; \textsc{FTR} captures premature or hallucinated actions; \textsc{TAR} reveals insufficient tool-calling capability or stalled conversational behaviors.

\paragraph{Precision and Recall Metrics.} As supplementary diagnostics, we compute precision and recall at both the tool-identifier and argument‐key levels.
\begin{itemize}[leftmargin=1.55em,itemsep=4pt]
  \item \textbf{Tool-Call Precision} (\textsc{TCP})
    \[
      \textsc{TCP}
      \;=\;
      \frac{\displaystyle
            \sum_{(c,g)\in\mathcal{M}}
              \bigl|\texttt{tnames}(c)\cap\texttt{tnames}(g)\bigr|}
           {\displaystyle
            \sum_{c\in C}
              \lvert\texttt{tnames}(c)\rvert}.
    \]
  \item \textbf{Tool-Call Recall} (\textsc{TCR})
    \[
      \textsc{TCR}
      \;=\;
      \frac{\displaystyle
            \sum_{(c,g)\in\mathcal{M}}
              \bigl|\texttt{tnames}(c)\cap\texttt{tnames}(g)\bigr|}
           {\displaystyle
            \sum_{g\in G}
              \lvert\texttt{tnames}(g)\rvert}.
    \]
  \item \textbf{Param-Key Precision} (\textsc{PKP})
    \[
      \textsc{PKP}
      \;=\;
      \frac{\displaystyle
            \sum_{(c,g)\in\mathcal{M}}
              \bigl|\texttt{keys}(c)\cap\texttt{keys}(g)\bigr|}
           {\displaystyle
            \sum_{c\in C}
              \lvert\texttt{keys}(c)\rvert}.
    \]
  \item \textbf{Param-Key Recall} (\textsc{PKR})
    \[
      \textsc{PKR}
      \;=\;
      \frac{\displaystyle
            \sum_{(c,g)\in\mathcal{M}}
              \bigl|\texttt{keys}(c)\cap\texttt{keys}(g)\bigr|}
           {\displaystyle
            \sum_{g\in G}
              \lvert\texttt{keys}(g)\rvert}.
    \]
\end{itemize}
\textsc{TCP} and \textsc{PKP} capture \emph{precision}, the fraction of predicted items that are correct, while \textsc{TCR} and \textsc{PKR} measure \emph{recall}, the fraction of reference items successfully included in the predicted items.  All four metrics lie in \([0,1]\), with higher values indicating better performance.

\paragraph{Conversational Quality Metrics.}
\label{secpara:conv_metrics}
While tool call correctness is paramount, an enterprise assistant must also sustain a clear, coherent, and diverse dialogue.  We therefore complement the tool-oriented scores (\textsc{Acc}, \textsc{FTR}, \textsc{TAR}) with three linguistic metrics that probe turn-level coherence and corpus-level lexical breadth.  
Unless otherwise noted, all computations exclude the assistant’s private \textit{thought} traces and consider only user-visible tokens produced by the model.

\begin{itemize}[leftmargin=1.4em,itemsep=3pt]
  \item \textbf{Conversation Relevancy} (\textsc{ConvRel}).  
        For each assistant reply \(a_{t}\) we query a \emph{rubric LLM} that judges how well the utterance builds on the dialogue prefix visible to the assistant,
        \(\mathbf{h}^{a}_{t}=(u_{1},a_{1},\dots,u_{t})\).
        The rubric emits an ordinal score
        \(s_{t}\in\{1,2,3\}\)
        (1 = off-topic, 2 = partly relevant, 3 = fully grounded). We map these raw grades to a normalized similarity
        \(\mathrm{sim}(a_{t},\mathbf{h}^{a}_{t}) \in \{0,0.5,1\}\)
        via
        \(g(1)=0,\;g(2)=0.5,\;g(3)=1\).
        Averaging over the \(T\) assistant turns of a dialogue
        \(d\) yields
        \[
          \textsc{ConvRel}(d)
          \;=\;
          \frac{1}{T}
          \sum_{t=1}^{T}      \mathrm{sim}\bigl(a_{t},\mathbf{h}^{a}_{t}\bigr).
        \]

  \item \textbf{Type–Token Ratio} (\textsc{TTR}).  
        Corpus-level lexical richness is measured by
        \[
          \textsc{TTR}(\mathcal{S})
          \;=\;
          \frac{\lvert\text{unique-1gram}(\mathcal{S})\rvert}
               {\lvert\text{all-1gram}(\mathcal{S})\rvert},
        \]
        where $\lvert\text{unique-1gram}(\mathcal{S})\rvert$ counts distinct surface word forms, and $\lvert\text{all-1gram}(\mathcal{S})\rvert$ denotes the total number of tokens in $\mathcal{S}$.

  \item \textbf{$n$-Gram Diversity} (\textsc{NGD}$_n$). 
        To capture syntactic variety beyond unigram choice, we compute the proportion of unique $n$-grams (here $n\!\in\!\{2,3,4\}$) relative to corpus length:
        \[
          \textsc{NGD}_{n}(\mathcal{S})
          \;=\;
          \frac{\lvert\text{unique-$n$gram}(\mathcal{S})\rvert}
               {\lvert\text{all-$n$gram}(\mathcal{S})\rvert}.
        \]
        Higher values indicate a broader repertoire of multi-word patterns and reduce the risk of template-like repetition.
\end{itemize}

\noindent
For all linguistic metrics, \textit{higher} is better.  When reported
together with \textsc{Acc}, \textsc{FTR}, and \textsc{TAR}, they offer
a holistic view: an ideal assistant both executes the right tools and
maintains engaging, contextually grounded prose.


\begin{table*}[htbp]
  \centering
  \small

  \setlength{\tabcolsep}{4pt}
  \begin{tabular}{l | ccccc cc}
    \hline
    \textbf{Model} & \textsc{TCP ($\uparrow$)} & \textsc{TCR ($\uparrow$)} & \textsc{PKP ($\uparrow$)} & \textsc{PKR ($\uparrow$)} & \textsc{Acc ($\uparrow$)} & \textsc{FTR ($\downarrow$)} & \textsc{TAR ($\downarrow$)} \\
    \hline
    Llama-3.2-DiaFORGE-3B & $\num{0.58}$ & $\num{0.58}$ & $\num{0.58}$ & $\num{0.57}$ & $\num{0.52}$ & $\num{0.12}$ & $\num{0.30}$ \\
    Llama-3.3-70B & $\num{0.03}$ & $\num{0.03}$ & $\num{0.03}$ & $\num{0.03}$ & $\num{0.03}$ & $\textbf{\num{0.00}}$ & $\num{0.97}$ \\
    Llama-3.3-70B-fc & $\num{0.47}$ & $\num{0.47}$ & $\num{0.47}$ & $\num{0.46}$ & $\num{0.22}$ & $\num{0.52}$ & $\textbf{\num{0.01}}$ \\
    Llama-3.3-DiaFORGE-70B & $\num{0.43}$ & $\num{0.44}$ & $\num{0.44}$ & $\num{0.44}$ & $\num{0.42}$ & $\num{0.03}$ & $\num{0.55}$ \\
    Llama-xLAM-2-70B-fc-r & $\num{0.72}$ & $\num{0.73}$ & $\num{0.73}$ & $\num{0.73}$ & $\num{0.48}$ & $\num{0.18}$ & $\num{0.13}$ \\
    Llama-3.3-Nemotron-Super-49B & $\num{0.68}$ & $\num{0.69}$ & $\num{0.69}$ & $\num{0.69}$ & $\num{0.60}$ & $\num{0.07}$ & $\num{0.25}$ \\
    Llama-3.3-Nemotron-DiaFORGE-49B & $\textbf{\num{0.84}}$ & $\textbf{\num{0.84}}$ & $\textbf{\num{0.84}}$ & $\textbf{\num{0.84}}$ & $\textbf{\num{0.82}}$ & $\num{0.04}$ & $\num{0.12}$ \\
    \hline
    Gemma-3-4B & $\num{0.25}$ & $\num{0.25}$ & $\num{0.25}$ & $\num{0.25}$ & $\num{0.19}$ & $\num{0.17}$ & $\num{0.61}$ \\
    Gemma-3-DiaFORGE-4B & $\num{0.58}$ & $\num{0.58}$ & $\num{0.58}$ & $\num{0.57}$ & $\num{0.53}$ & $\num{0.05}$ & $\num{0.37}$ \\
    Gemma-3-12B & $\num{0.34}$ & $\num{0.34}$ & $\num{0.34}$ & $\num{0.34}$ & $\num{0.31}$ & $\num{0.03}$ & $\num{0.62}$ \\
    Gemma-3-DiaFORGE-12B & $\num{0.72}$ & $\num{0.72}$ & $\num{0.72}$ & $\num{0.72}$ & $\num{0.68}$ & $\textbf{\num{0.02}}$ & $\num{0.26}$ \\
    Gemma-3-27B & $\num{0.20}$ & $\num{0.20}$ & $\num{0.20}$ & $\num{0.20}$ & $\num{0.19}$ & $\textbf{\num{0.02}}$ & $\num{0.78}$ \\
    Gemma-3-DiaFORGE-27B & $\textbf{\num{0.79}}$ & $\textbf{\num{0.79}}$ & $\textbf{\num{0.79}}$ & $\textbf{\num{0.79}}$ & $\textbf{\num{0.77}}$ & $\num{0.03}$ & $\textbf{\num{0.18}}$ \\
    \hline
    GPT-4o-20241120 & $\num{0.19}$ & $\num{0.19}$ & $\num{0.19}$ & $\num{0.19}$ & $\num{0.19}$ & $\num{0.00}$ & $\num{0.81}$ \\
    GPT-4o-20241120-fc & $\num{0.62}$ & $\num{0.82}$ & $\num{0.82}$ & $\num{0.81}$ & $\num{0.61}$ & $\num{0.64}$ & $\num{0.16}$ \\
    Claude-3.5-Sonnet-20241022 & $\num{0.17}$ & $\num{0.17}$ & $\num{0.17}$ & $\num{0.17}$ & $\num{0.15}$ & $\num{0.02}$ & $\num{0.82}$ \\
    Claude-3.5-Sonnet-20241022-fc & $\num{0.62}$ & $\num{0.76}$ & $\num{0.76}$ & $\num{0.76}$ & $\num{0.42}$ & $\num{0.76}$ & $\num{0.03}$ \\
    \hline
  \end{tabular}
  \caption{\label{tab:appendix_static_tool_result}Static Evaluation Results for Tool-Calling Metrics. Bootstrap 95\,\% CIs for \textsc{Acc} are reported in Table~\ref{tab:bootstrap_cis}.}
  \setlength{\tabcolsep}{6pt}
\end{table*}

\begin{table*}[htbp]
  \centering
  \small

  \setlength{\tabcolsep}{4pt}
  \begin{tabular}{l | ccc}
    \hline
    \textbf{Model} & \textsc{ConvRel ($\uparrow$)} & \textsc{TTR ($\uparrow$)} & \textsc{NGD$_3$ ($\uparrow$)} \\
    \hline
    Llama-3.2-DiaFORGE-3B & $\num{0.75}$ & $\num{0.13}$ & $\num{0.58}$ \\
    Llama-3.3-70B & $\num{0.95}$ & $\num{0.13}$ & $\num{0.61}$ \\
    Llama-3.3-70B-fc & $\num{0.43}$ & $\num{0.20}$ & $\num{0.30}$ \\
    Llama-3.3-DiaFORGE-70B & $\num{0.96}$ & $\num{0.10}$ & $\num{0.55}$ \\
    Llama-xLAM-2-70B-fc-r & $\num{0.73}$ & $\num{0.11}$ & $\num{0.58}$ \\
    Llama-3.3-Nemotron-Super-49B & $\num{0.74}$ & $\num{0.11}$ & $\num{0.60}$ \\
    Llama-3.3-Nemotron-DiaFORGE-49B & $\num{0.82}$ & $\num{0.14}$ & $\num{0.58}$ \\
    \hline
    Gemma-3-4B & $\num{0.72}$ & $\num{0.15}$ & $\num{0.56}$ \\
    Gemma-3-DiaFORGE-4B & $\num{0.81}$ & $\num{0.12}$ & $\num{0.57}$ \\
    Gemma-3-12B & $\num{0.75}$ & $\num{0.17}$ & $\num{0.64}$ \\
    Gemma-3-DiaFORGE-12B & $\num{0.82}$ & $\num{0.13}$ & $\num{0.57}$ \\
    Gemma-3-27B & $\num{0.95}$ & $\num{0.16}$ & $\num{0.66}$ \\
    Gemma-3-DiaFORGE-27B & $\num{0.84}$ & $\num{0.13}$ & $\num{0.57}$ \\
    \hline
    GPT-4o-20241120 & $\num{0.98}$ & $\num{0.16}$ & $\num{0.73}$ \\
    GPT-4o-20241120-fc & $\num{0.89}$ & $\num{0.10}$ & $\num{0.63}$ \\
    Claude-3.5-Sonnet-20241022 & $\num{0.93}$ & $\num{0.10}$ & $\num{0.58}$ \\
    Claude-3.5-Sonnet-20241022-fc & $\num{0.52}$ & $\num{0.12}$ & $\num{0.67}$ \\
    \hline
  \end{tabular}
  \caption{\label{tab:appendix_static_conv_result}Static Evaluation Results for Conversational Metrics}
  \setlength{\tabcolsep}{6pt}
\end{table*}

\begin{table*}[htbp]
  \centering
  \small

  \setlength{\tabcolsep}{4pt}
  \begin{tabular}{l | ccccc cc}
    \hline
    \textbf{Model} & \textsc{TCP ($\uparrow$)} & \textsc{TCR ($\uparrow$)} & \textsc{PKP ($\uparrow$)} & \textsc{PKR ($\uparrow$)} & \textsc{Acc ($\uparrow$)} & \textsc{FTR ($\downarrow$)} & \textsc{TAR ($\downarrow$)} \\
    \hline
    Llama-3.2-DiaFORGE-3B & $\num{0.86}$ & $\num{0.86}$ & $\num{0.86}$ & $\num{0.85}$ & $\num{0.80}$ & $\num{0.08}$ & $\num{0.06}$ \\
    Llama-3.3-70B & $\num{0.11}$ & $\num{0.11}$ & $\num{0.11}$ & $\num{0.11}$ & $\num{0.11}$ & $\textbf{\num{0.02}}$ & $\num{0.88}$ \\
    Llama-3.3-70B-fc & $\num{0.77}$ & $\num{0.77}$ & $\num{0.77}$ & $\num{0.77}$ & $\num{0.30}$ & $\num{0.22}$ & $\textbf{\num{0.01}}$ \\
    Llama-3.3-DiaFORGE-70B & $\num{0.77}$ & $\num{0.77}$ & $\num{0.77}$ & $\num{0.77}$ & $\num{0.71}$ & $\num{0.04}$ & $\num{0.19}$ \\
    Llama-xLAM-2-70B-fc-r & $\num{0.87}$ & $\num{0.89}$ & $\num{0.89}$ & $\num{0.89}$ & $\num{0.51}$ & $\num{0.18}$ & $\num{0.05}$ \\
    Llama-3.3-Nemotron-Super-49B & $\num{0.86}$ & $\num{0.87}$ & $\num{0.87}$ & $\num{0.86}$ & $\num{0.72}$ & $\num{0.08}$ & $\num{0.08}$ \\
    Llama-3.3-Nemotron-DiaFORGE-49B & $\textbf{\num{0.92}}$ & $\textbf{\num{0.92}}$ & $\textbf{\num{0.92}}$ & $\textbf{\num{0.92}}$ & $\textbf{\num{0.89}}$ & $\num{0.06}$ & $\num{0.03}$ \\
    \hline
    Gemma-3-4B & $\num{0.32}$ & $\num{0.32}$ & $\num{0.32}$ & $\num{0.31}$ & $\num{0.24}$ & $\num{0.14}$ & $\num{0.58}$ \\
    Gemma-3-DiaFORGE-4B & $\num{0.86}$ & $\num{0.86}$ & $\num{0.86}$ & $\num{0.85}$ & $\num{0.81}$ & $\num{0.09}$ & $\num{0.05}$ \\
    Gemma-3-12B & $\num{0.39}$ & $\num{0.39}$ & $\num{0.39}$ & $\num{0.39}$ & $\num{0.37}$ & $\num{0.04}$ & $\num{0.57}$ \\
    Gemma-3-DiaFORGE-12B & $\num{0.87}$ & $\num{0.87}$ & $\num{0.87}$ & $\num{0.87}$ & $\num{0.86}$ & $\num{0.07}$ & $\num{0.07}$ \\
    Gemma-3-27B & $\num{0.21}$ & $\num{0.21}$ & $\num{0.21}$ & $\num{0.21}$ & $\num{0.21}$ & $\textbf{\num{0.00}}$ & $\num{0.79}$ \\
    Gemma-3-DiaFORGE-27B & $\textbf{\num{0.94}}$ & $\textbf{\num{0.94}}$ & $\textbf{\num{0.94}}$ & $\textbf{\num{0.94}}$ & $\textbf{\num{0.89}}$ & $\num{0.03}$ & $\textbf{\num{0.03}}$ \\
    \hline
    GPT-4o-20241120 & $\num{0.63}$ & $\num{0.63}$ & $\num{0.63}$ & $\num{0.63}$ & $\num{0.62}$ & $\num{0.02}$ & $\num{0.36}$ \\
    GPT-4o-20241120-fc & $\num{0.74}$ & $\num{0.87}$ & $\num{0.87}$ & $\num{0.87}$ & $\num{0.56}$ & $\num{0.59}$ & $\num{0.05}$ \\
    Claude-3.5-Sonnet-20241022 & $\num{0.43}$ & $\num{0.43}$ & $\num{0.43}$ & $\num{0.43}$ & $\num{0.39}$ & $\num{0.03}$ & $\num{0.55}$ \\
    Claude-3.5-Sonnet-20241022-fc & $\num{0.76}$ & $\num{0.82}$ & $\num{0.82}$ & $\num{0.82}$ & $\num{0.40}$ & $\num{0.34}$ & $\num{0.03}$ \\
    \hline
  \end{tabular}
  \caption{\label{tab:appendix_dynamic_tool_result}Dynamic Evaluation Results for Tool-Calling Metrics. Bootstrap 95\,\% CIs for \textsc{Acc} are reported in Table~\ref{tab:bootstrap_cis}.}
  \setlength{\tabcolsep}{6pt}
\end{table*}

\begin{table*}[htbp]
  \centering
  \small

  \setlength{\tabcolsep}{4pt}
  \begin{tabular}{l | ccc}
    \hline
    \textbf{Model} & \textsc{ConvRel ($\uparrow$)} & \textsc{TTR ($\uparrow$)} & \textsc{NGD$_3$ ($\uparrow$)} \\
    \hline
    Llama-3.2-DiaFORGE-3B & $\num{0.80}$ & $\num{0.13}$ & $\num{0.54}$ \\
    Llama-3.3-70B & $\num{0.94}$ & $\num{0.14}$ & $\num{0.60}$ \\
    Llama-3.3-70B-fc & $\num{0.54}$ & $\num{0.11}$ & $\num{0.20}$ \\
    Llama-3.3-DiaFORGE-70B & $\num{0.94}$ & $\num{0.11}$ & $\num{0.58}$ \\
    Llama-xLAM-2-70B-fc-r & $\num{0.69}$ & $\num{0.11}$ & $\num{0.48}$ \\
    Llama-3.3-Nemotron-Super-49B & $\num{0.73}$ & $\num{0.10}$ & $\num{0.51}$ \\
    Llama-3.3-Nemotron-DiaFORGE-49B & $\num{0.85}$ & $\num{0.13}$ & $\num{0.54}$ \\
    \hline
    Gemma-3-4B & $\num{0.70}$ & $\num{0.17}$ & $\num{0.58}$ \\
    Gemma-3-DiaFORGE-4B & $\num{0.84}$ & $\num{0.13}$ & $\num{0.57}$ \\
    Gemma-3-12B & $\num{0.77}$ & $\num{0.19}$ & $\num{0.62}$ \\
    Gemma-3-DiaFORGE-12B & $\num{0.84}$ & $\num{0.13}$ & $\num{0.54}$ \\
    Gemma-3-27B & $\num{0.91}$ & $\num{0.18}$ & $\num{0.67}$ \\
    Gemma-3-DiaFORGE-27B & $\num{0.85}$ & $\num{0.15}$ & $\num{0.61}$ \\
    \hline
    GPT-4o-20241120 & $\num{0.93}$ & $\num{0.15}$ & $\num{0.69}$ \\
    GPT-4o-20241120-fc & $\num{0.69}$ & $\num{0.07}$ & $\num{0.43}$ \\
    Claude-3.5-Sonnet-20241022 & $\num{0.92}$ & $\num{0.09}$ & $\num{0.54}$ \\
    Claude-3.5-Sonnet-20241022-fc & $\num{0.52}$ & $\num{0.07}$ & $\num{0.46}$ \\
    \hline
  \end{tabular}
  \caption{\label{tab:appendix_dynamic_conv_result}Dynamic Evaluation Results for Conversational Metrics}
  \setlength{\tabcolsep}{6pt}
\end{table*}

\subsection{Computational Results}
\label{sec:appendix_C2}

We evaluate all models listed in Table~\ref{tab:tool_eval} using both the static and dynamic metrics described in Appendix~\ref{sec:appendix_C1}.

\paragraph{Results for Tool-Calling Metrics.} Static results are given in Table~\ref{tab:appendix_static_tool_result}, and dynamic results in Table~\ref{tab:appendix_dynamic_tool_result}. DiaFORGE fine-tuning consistently boosts performance across all Llama-3 and Gemma-3 backbones. The strongest models are \textit{Llama-3.3-Nemotron-DiaFORGE-49B} and \textit{Gemma-3-DiaFORGE-27B}, each of which substantially outperforms GPT-4o and Claude-3.5-Sonnet. Model size is not a monotonic indicator of quality: the compact \textit{Llama-3.2-DiaFORGE-3B} and the mid-sized \textit{Llama-3.3-Nemotron-DiaFORGE-49B} both surpass the much larger \textit{Llama-3.3-DiaFORGE-70B}.

\paragraph{Per-model bootstrap 95\,\% CIs.}
Table~\ref{tab:bootstrap_cis} reports bootstrap 95\,\% CIs for both static and dynamic \textsc{Acc} for all 17 evaluated models.

\begin{table*}[h]
  \centering
  \small
  \setlength{\tabcolsep}{5pt}
  \begin{tabular}{l cc cc}
    \toprule
    & \multicolumn{2}{c}{\textbf{Static Acc}} & \multicolumn{2}{c}{\textbf{Dynamic Acc}} \\
    \cmidrule(lr){2-3}\cmidrule(lr){4-5}
    \textbf{Model} & \textbf{Acc} & \textbf{95\,\% CI} & \textbf{Acc} & \textbf{95\,\% CI} \\
    \midrule
    Llama-3.2-DiaFORGE-3B         & 0.52 & [0.43, 0.61] & 0.80 & [0.73, 0.87] \\
    Llama-3.3-70B                  & 0.03 & [0.00, 0.08] & 0.11 & [0.06, 0.17] \\
    Llama-3.3-70B-fc               & 0.22 & [0.15, 0.29] & 0.30 & [0.22, 0.38] \\
    Llama-3.3-DiaFORGE-70B        & 0.42 & [0.33, 0.52] & 0.79 & [0.71, 0.87] \\
    Llama-xLAM-2-70B-fc-r         & 0.48 & [0.40, 0.57] & 0.51 & [0.43, 0.60] \\
    Llama-3.3-Nemotron-Super-49B  & 0.60 & [0.51, 0.69] & 0.72 & [0.64, 0.80] \\
    Llama-3.3-Nemotron-DiaFORGE-49B & \textbf{0.82} & [\textbf{0.76, 0.89}] & \textbf{0.89} & [\textbf{0.83, 0.94}] \\
    \hline
    Gemma-3-4B                     & 0.19 & [0.12, 0.26] & 0.24 & [0.17, 0.32] \\
    Gemma-3-DiaFORGE-4B           & 0.53 & [0.44, 0.62] & 0.81 & [0.74, 0.87] \\
    Gemma-3-12B                    & 0.31 & [0.23, 0.39] & 0.37 & [0.28, 0.45] \\
    Gemma-3-DiaFORGE-12B          & 0.68 & [0.60, 0.76] & 0.86 & [0.80, 0.92] \\
    Gemma-3-27B                    & 0.19 & [0.12, 0.26] & 0.21 & [0.14, 0.29] \\
    Gemma-3-DiaFORGE-27B          & \textbf{0.77} & [\textbf{0.70, 0.85}] & \textbf{0.89} & [\textbf{0.83, 0.94}] \\
    \hline
    GPT-4o-20241120                & 0.19 & [0.12, 0.26] & 0.62 & [0.54, 0.70] \\
    GPT-4o-20241120-fc             & 0.61 & [0.52, 0.70] & 0.56 & [0.47, 0.65] \\
    Claude-3.5-Sonnet-20241022     & 0.15 & [0.09, 0.22] & 0.39 & [0.30, 0.48] \\
    Claude-3.5-Sonnet-20241022-fc  & 0.42 & [0.33, 0.51] & 0.40 & [0.31, 0.48] \\
    \bottomrule
  \end{tabular}
  \caption{Bootstrap 95\,\% CIs for static and dynamic \textsc{Acc} across all 17 models ($N{=}119$ test cases, $N{=}2000$ resamples, percentile method). The CI width reflects the small test-set size common to proprietary enterprise benchmarks.}
  \label{tab:bootstrap_cis}
\end{table*}

\paragraph{Results for Conversational Metrics.} Table~\ref{tab:appendix_static_conv_result} and Table~\ref{tab:appendix_dynamic_conv_result} report the results of static and dynamic conversational evaluations, respectively. These metrics are intended to verify that fine-tuning preserves general dialogue competence. Since \textsc{ConvRel} is computed using an LLM-based evaluator, its values should be interpreted as heuristic estimates rather than precise measurements. Our primary goal is to assess the relative conversational relevance of fine-tuned models compared to their instruction-tuned baselines and proprietary models such as GPT-4o and Claude-3.5-Sonnet. Across all backbone models, DiaFORGE fine-tuning maintains conversational quality, showing no statistically significant degradation while often matching or surpassing the performance of proprietary counterparts.

\subsection{Multi-Sampling Voting Mechanism of User-Proxy in Dynamic Evaluation}
\label{sec:appendix_C3}

\begin{figure}[h]
\centering
\includegraphics[width=\columnwidth, trim = 3cm 0cm 3cm 0cm, clip]{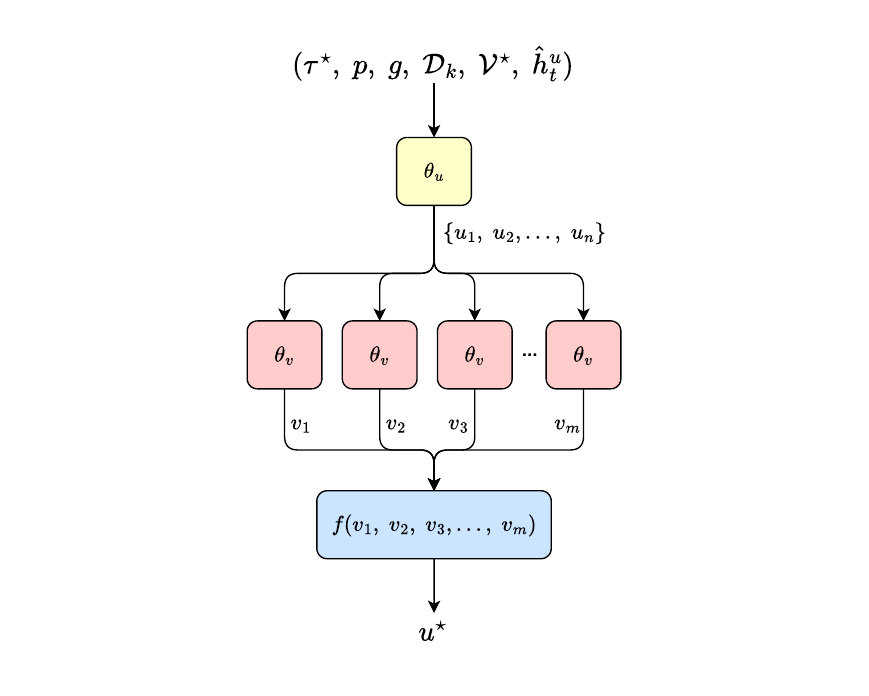}
\caption{Reducing hallucination for user utterance generation in dynamic evaluation by applying a multi-sampling and voting strategy.}
\label{fig:dynamic_user}
\end{figure}

\begin{table*}[htbp]
  \centering
  \begin{tabular}{l | ccc}
    \hline
    \textbf{Model ($\theta_u$)} & \textsc{Acc ($\uparrow$)} & \textsc{FTR ($\downarrow$)} & \textsc{TAR ($\downarrow$)} \\
    \hline
    GPT-4o-20241120 & $\num{0.8861}$ & $\num{0.0588}$ & $\num{0.0252}$ \\
    Claude-3.5-Sonnet-20241022 & $\num{0.8627}$ & $\num{0.0957}$ & $\num{0.0000}$ \\
    Llama-3.1-70B & $\num{0.8897}$ & $\num{0.0420}$ & $\num{0.0168}$ \\
    Mistral-Large-1124 & $\num{0.8768}$ & $\num{0.0504}$ & $\num{0.0336}$ \\\hline
  \end{tabular}
  \caption{Effect of varying the user-proxy model $\theta_u$ on user utterance generation in dynamic evaluation, with \textit{Llama-3.3-Nemotron-DiaFORGE-49B} fixed as the assistant agent.}
  \label{tab:diff_models_for_userproxy_results}
\end{table*}

Dynamic evaluation differs from static evaluation primarily in how user utterances are generated. While static evaluation reuses pre-generated user inputs, dynamic evaluation generates user utterances adaptively based on the current chat history \(\hat{\mathbf{h}}_t^u\).  As detailed in §\ref{sec:utc_gen}, the user-proxy LLM parameterized by \(\boldsymbol{\theta}_u\) is responsible for generating user utterances conditioned on a structured context tuple:
\[
  (\tau^\star,\; p,\; g,\; \mathcal{D}_k,\; \mathcal{V}^\star,\; \hat{\mathbf{h}}_t^u).
\]
During synthetic dialogue generation, hallucinations are filtered post-hoc via a validation stage. However, dynamic evaluation forgoes rejection-based filtering to preserve evaluation coverage. The validation mechanism described in §\ref{par:validator} does not exempt user-proxy hallucinations, whereas dynamic evaluation is intended to assess only assistant agent performance. Any hallucination originating from the user-proxy introduces noise and undermines this evaluation goal.

To address this, we introduce a multi-sampling and voting scheme to stabilize user utterance generation, illustrated in Figure~\ref{fig:dynamic_user}. The method leverages two distinct LLMs: a generator LLM with parameters \(\boldsymbol{\theta}_u\), and a voter LLM with parameters \(\boldsymbol{\theta}_v\).

We begin by independently sampling a set of \(n\) candidate utterances from the generator:
\[
  U = \left\{ u_i \sim P_{\boldsymbol{\theta}_u} \left( \cdot \;\middle|\; \tau^\star,\; p,\; g,\; \mathcal{D}_k,\; \mathcal{V}^\star,\; \hat{\mathbf{h}}_t^u \right) \right\}_{i=1}^{n}.
\]

Next, the \(n\) utterances in \(U\) are evaluated by \(m\) independent voters, each instantiated with \(\boldsymbol{\theta}_v\). Each voter is tasked with selecting the \emph{single best} candidate utterance from the set \(U = \{u_1, \dots, u_n\}\). To reduce positional bias, the utterances are randomly permuted prior to presentation. For each voter \(j = 1, \dots, m\), let \(\pi_j : [n] \rightarrow [n]\) denote the permutation applied to the indices. The vote is then drawn as:
\[
  v_j \sim P_{\boldsymbol{\theta}_v} \left( \{1, \dots, n\} \;\middle|\; p,\; g,\ \hat{\mathbf{h}}_t^u,\; \pi_j(U) \right),
\]
where \(v_j \in \{1, \dots, n\}\) denotes the index of the utterance selected from the permuted list \(\pi_j(U)\). We then invert the permutation to recover the index with respect to the original candidate set \(U\).

Finally, the votes \(\{ v_1, \dots, v_m \}\) are aggregated via a deterministic pooling function
\[
  f: \{1, \dots, n\}^m \rightarrow \{1, \dots, n\},
\]
typically instantiated as the mode operator. The final user utterance is selected as:
\[
  u^\star = u_{f(v_1,\; \dots,\; v_m)}.
\]

In the dynamic evaluation results presented in Table~\ref{tab:tool_eval}, both the generator model $\boldsymbol{\theta}_u$ and the voter model $\boldsymbol{\theta}_v$ are configured with GPT-4o. We use a sampling size of $n = 3$, a voting ensemble of $m = 3$, and apply mode pooling as the aggregation function $f$.

To assess the sensitivity of dynamic evaluation outcomes to the choice of $\theta_u$, we conduct ablation experiments in which the fine-tuned assistant model \textit{Llama-3.3-Nemotron-DiaFORGE-49B} is paired with various alternative user-proxy models. The results are summarized in Table~\ref{tab:diff_models_for_userproxy_results}, with all other hyperparameters held fixed.

Across configurations, we observe only minor fluctuations in the evaluation metrics. A closer inspection of the divergent cases reveals that the hallucinations predominantly originate from the assistant model itself. Moreover, because dynamic evaluation permits the assistant to explore multiple plausible dialogue trajectories, small variations (on the order of a few percentage points) are expected and not indicative of true performance shifts.  As such, comparisons between assistant models under dynamic evaluation are only meaningful when the observed performance differences are sufficiently large to outweigh inherent evaluation variance.

\subsection{Choice of Different LLMs}

In this study, we intentionally excluded certain models. For example, although Mistral models are among the leading open-source options, we did not include them due to their non-standard and heterogeneous chat template formatting, which complicates consistent evaluation.

Additionally, we omit baseline results for the \textit{Llama-3.2-3B-Instruct} model, as it exhibited near-zero performance on the tool-calling metrics.

\begin{table*}[t]
  \centering
  \scriptsize
  \setlength{\tabcolsep}{2pt}
  \begin{tabular}{lcccc}
    \toprule
    \textbf{\textsc{BFCL v3} Metric} &
    \textbf{Gemma-3-27B (FC)} &
    \textbf{Gemma-3-DiaFORGE-27B (FC)} &
    \textbf{Gemma-3-27B (prompt)} &
    \textbf{Gemma-3-DiaFORGE-27B (prompt)} \\
    \midrule
    Single-Turn Non-Live AST Acc & 89.19\% & 89.17\% & 89.19\% & 89.29\% \\
    Single-Turn Live Acc         & 76.19\% & 76.14\% & 76.10\% & 75.97\% \\
    Multi-Turn Acc               & 15.00\% & 14.88\% & 14.88\% & 15.00\% \\
    Relevance Detection          & 83.33\% & 83.33\% & 83.33\% & 83.33\% \\
    Irrelevance Detection        & 73.10\% & 73.10\% & 73.05\% & 72.88\% \\
    \textbf{Overall Acc}         & \textbf{59.26\%} & \textbf{59.20\%} & \textbf{59.19\%} & \textbf{59.22\%} \\
    \bottomrule
  \end{tabular}
  \caption{\textbf{Gemma-3-27B} vs.\ \textbf{Gemma-3-DiaFORGE-27B} on \textsc{BFCL v3}: scores are essentially unchanged under both FC and ``prompt'', indicating parity (no overfitting). BFCL queries are fully specified and do not probe disambiguation.}
  \label{tab:bfcl_g27b}
\end{table*}

\subsection{Parity Check on Public Benchmarks}
\label{sec:bfcl}

In this section, we evaluate and compare the performance of the \textsc{DiaFORGE}-tuned model with the base model in order to verify any trace of overfitting or catastrophic forgetting.

As discussed in Appendix~\ref{sec:comp_diabench_and_bfcl}, the general-purpose \textsc{BFCL v3} benchmark provides minimal coverage of the disambiguation behaviors targeted by \textsc{DiaFORGE}. Nonetheless, to rule out overfitting, we evaluate \emph{Gemma-3-27B} and \emph{Gemma-3-DiaFORGE-27B} on \textsc{BFCL v3}~\citep{yan2024bfcl}. Because \textsc{BFCL} requests are largely fully specified, with few near-duplicate tools and little missing-argument pressure, our objective is to demonstrate \emph{parity} with the base model rather than gains. We report results for both prompt-based (prompt) and native function calling (FC), and we additionally compute a pairwise win rate of \emph{Gemma-3-DiaFORGE-27B} versus \emph{Gemma-3-27B} on MT-Bench ~\citep{zheng2023judging} to check for any regressions in general model capability attributable to disambiguation-centric fine-tuning.

Performance on \textsc{BFCL v3} remains essentially unchanged across all metrics (Table~\ref{tab:bfcl_g27b}), indicating no overfitting to our data and no catastrophic forgetting of native function calling skills. On \textsc{MT-Bench}/Chatbot Arena with GPT-4o as judge, \textsc{Gemma-3-DiaFORGE-27B} attains a pairwise win rate of \textbf{0.50} against \textsc{Gemma-3-27B}, suggesting parity on general model capabilities rather than degradation.

\subsection{Disambiguation Frontier: Details and Stratified Analysis}
\label{sec:appendix_disamb_frontier}

\paragraph{Similarity computation.}
For each of the \textsc{DiaBENCH} test cases, we embed the ground-truth target tool and every distractor tool using the same \texttt{text-embedding-3-large} encoder as described in Appendix~\ref{sec:appendix_A4}. The closest-distractor similarity is defined as the maximum cosine similarity between the target tool embedding and any distractor tool embedding.

\paragraph{Stratified breakdown.}
Table~\ref{tab:stratified_frontier} reports dynamic \textsc{Acc} for five representative models across a $3\times 2$ grid of similarity tier (low: $<\!0.65$, medium: $0.65$--$0.80$, high: $\geq\!0.80$) and dialogue length (1--2 turns vs.\ 3+ turns).

\begin{table*}[h]
  \centering
  \small
  \setlength{\tabcolsep}{4pt}
  \begin{tabular}{l | cc | cc | cc}
    \toprule
    & \multicolumn{2}{c|}{\textbf{Low sim} ($<\!0.65$)}
    & \multicolumn{2}{c|}{\textbf{Med sim} ($0.65$--$0.80$)}
    & \multicolumn{2}{c}{\textbf{High sim} ($\geq\!0.80$)} \\
    \textbf{Model} & 1--2 & 3{+} & 1--2 & 3{+} & 1--2 & 3{+} \\
    \midrule
    Gemma-3-DiaFORGE-27B  & 0.877 & 0.914 & 0.909 & 0.926 & 0.890 & 0.727 \\
    Gemma-3-27B           & 0.211 & 0.143 & 0.333 & 0.059 & 0.240 & 0.091 \\
    Nemotron-DiaFORGE-49B & 0.842 & 0.870 & 0.939 & 0.957 & 0.840 & 0.818 \\
    GPT-4o-20241120       & 0.579 & 0.500 & 0.818 & 0.569 & 0.720 & 0.091 \\
    Claude-3.5-Sonnet     & 0.368 & 0.286 & 0.606 & 0.157 & 0.400 & 0.273 \\
    \bottomrule
  \end{tabular}
  \caption{Dynamic \textsc{Acc} stratified by distractor similarity tier and dialogue turn count. \textsc{DiaFORGE}-tuned models exhibit minimal degradation across all conditions; closed-source and untuned baselines show sharp drops at higher similarity and longer dialogues.}
  \label{tab:stratified_frontier}
\end{table*}

\section{Production Case Study}
\label{sec:prod_case_study}

We present a production case study that, from a user-experience perspective, illustrates how disambiguation-aware LLM behavior reduces user friction. Consider the following persona: a newly hired team manager who needs help approving internal training requests from team members.

\begin{figure}[ht]
\centering
\includegraphics[width=\columnwidth, trim = 1cm 0.5cm 0.5cm 0.5cm, clip]{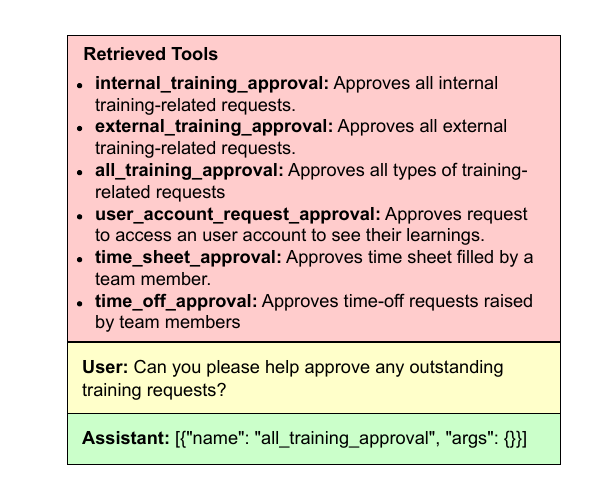}
\caption{Conversation between a real user and a closed-source model with native function calling.}
\label{fig:gpt_4o_manager_session}
\end{figure}

\begin{figure}[ht]
\centering
\includegraphics[width=\columnwidth, trim = 1cm 0.5cm 0.5cm 0.5cm, clip]{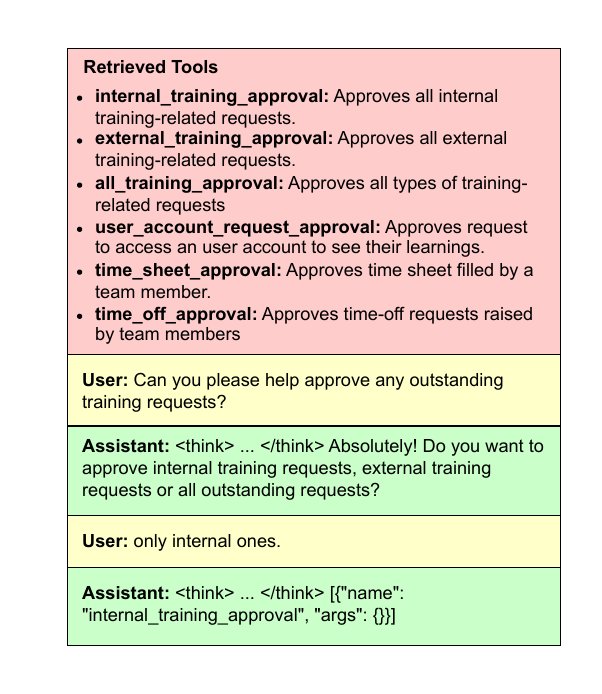}
\caption{Conversation between a real user and a \textsc{DiaFORGE}-fine-tuned model.}
\label{fig:gemma-3-diaforge-chat-session}
\end{figure}

We compare two models: (1) a closed-source model with native function calling, and (2) a \textsc{DiaFORGE}-tuned model. To ensure comparability, we hold fixed the initial user query and the retrieved tool set. Both models must converse with the user to elicit requirements and then issue a tool call.

Figure~\ref{fig:gpt_4o_manager_session} shows a conversation where the user asks to approve any outstanding training requests. Among the retrieved tools, three near-duplicate candidates are plausible targets: \textsc{internal\_training\_approval}, \textsc{external\_training\_approval}, and \textsc{all\_training\_approval}. Given the user persona, the intended action is to approve \emph{internal} requests only; approving external requests would draw from the team's budget. Without clarification, the closed-source model directly calls \textsc{all\_training\_approval}.

Figure~\ref{fig:gemma-3-diaforge-chat-session} illustrates the \textsc{DiaFORGE}-fine-tuned model: it first poses a targeted clarifying question to determine the request's scope (internal vs.\ external vs.\ all), then invokes the correct tool, \textsc{internal\_training\_approval}.

In production, guardrails could display a confirmation dialog before executing tool calls that perform write operations or incur costs, giving the user the final say to accept or cancel. However, issuing an overly broad or incorrect tool call without first clarifying the user's intent still creates friction. In the scenario of Figure~\ref{fig:gpt_4o_manager_session}, the user would see a cost warning, likely cancel, and then need to restate their requirements in greater detail, adding unnecessary back-and-forth.

Repeated occurrences of such misfires nudge users to over-specify queries up front, diminishing conversational naturalness and dampening engagement with the AI system. Over time, this friction erodes usage and, ultimately, market capitalization.

\section{System Prompt Optimization}
\label{sec:appendix_E}

As discussed in §\ref{sec:experiments}, we employ the Cost-Aware Prompt Optimization (CAPO) strategy to adapt system prompts for all evaluated models, leveraging their generation capabilities.

\begin{figure}[ht]
  \centering
  \includegraphics[width=\columnwidth]{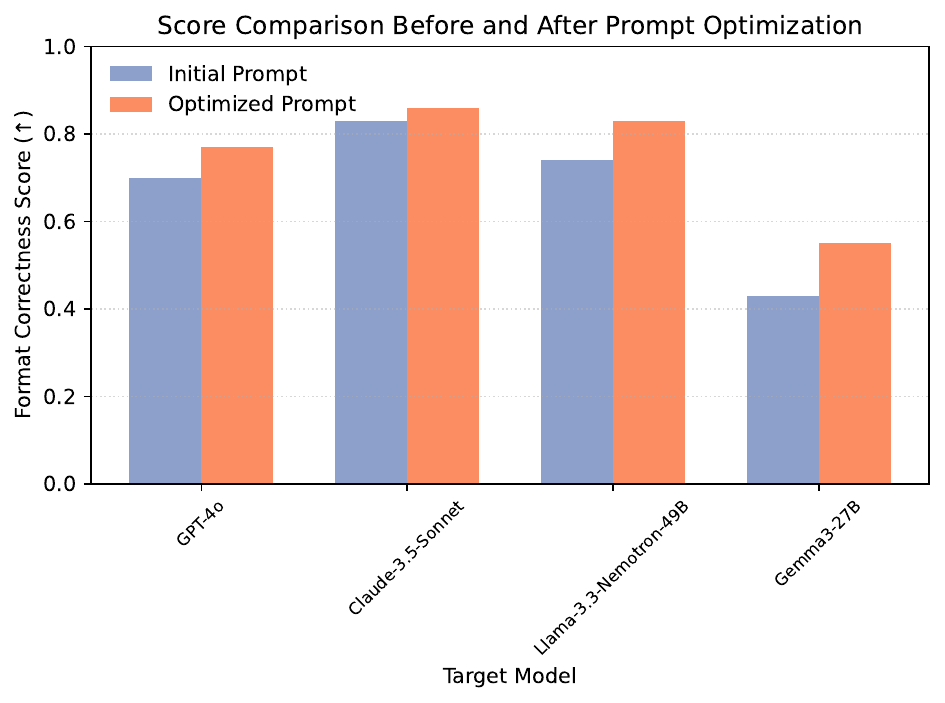}
  \caption{Format correctness score of various LLMs on the holdout set before and after prompt optimization.}
  \label{fig:capo_scores}
\end{figure}

The CAPO algorithm is parameterized as follows: significance level $\alpha = 0.2$ for the paired $t$-test used in racing; block size $b = 30$, indicating the number of development examples evaluated per batch; maximum number of blocks before discarding a candidate $z_{\text{max}} = 10$; upper bound on few-shot examples injected into a prompt $k_{\text{max}} = 5$; number of retained candidates per generation $\mu = 10$; number of crossovers per iteration $c = 4$; length penalty $\gamma = 0.05$; and maximum number of iterations $T = 10$. Each optimization run is given an unlimited token budget.

After each CAPO iteration, we evaluate the candidate prompts on the holdout set using the \textsc{Format Correctness} metric.  Figure~\ref{fig:capo_scores} presents the \textsc{Format Correctness} scores attained by the best-performing optimized prompts for each evaluated LLM.

We use a standardized reference system prompt to evaluate each fine-tuned model, which also serves as the initial input to CAPO. Each model is optimized using its own architecture unless it is a downstream variant of a base model, in which case we reuse the optimized prompt from the base model. For anonymization, all organization names in prompts are replaced with the placeholder \textsc{XYZ}.

Below, we provide the reference system prompt, along with examples of CAPO-optimized prompts for the following model families: \textit{GPT-4o}, \textit{Claude-3.5-Sonnet}, \textit{LLaMA-3.3}, and \textit{Gemma-3}.
 
\begin{tcolorbox}[enhanced jigsaw,breakable,pad at break*=1mm,
  colback=blue!5!white,colframe=blue!75!black,title=Initial Reference System Prompt]
 ===== Instructions =====  
 
You are an **AI assistant created by XYZ**.  
Your job unfolds in **two consecutive phases**:

---

\#\#\#\# Phase 1 - Tool Selection  

1. Review the list in **“Available Tools”**.  

2. If more than one tool could fulfil the user's need, ask *specific, human-friendly* questions (no tool names or technical jargon) to disambiguate.  

3. Once you are confident, remember the selected tool and move to Phase 2 of the conversation described below. Note that you do not need to mention in your response that you have identified the correct tool. Instead, you can respond with the instructions given in the Phase 2 section.  \\

\#\#\#\# Phase 2 - Parameter Collection \& Final Tool Call  

1. With the chosen tool identified, collect any missing parameters:  
   - Skip parameters the user has already provided.  
   - Ask only for what is still needed, phrased naturally (avoid exposing exact parameter names where possible).  

2. When all required parameters are gathered (optional ones may be omitted if not discussed), build a list of tool calls entries where each entry includes:  
   - `name`: chosen tool name  
   - `args`: JSON object containing every collected parameter/value  

3. Respond with this list containing tool calls (an empty `"args": \{\}` if the selected tool does not have any input parameters).  

4. Whenever you raise a tool call (list containing toolcalls), there should be empty and the response (other than thought between <think> </think>) should only be list containing toolcalls.

---

==== General Guidelines ====  

1. **Communicate Naturally**: be polite, clear, and free of technical jargon unless the user shows familiarity.  

2. **Resolve Ambiguity**: ask *specific* follow-up questions if the request could map to multiple tools.  

3. **Completeness**:  
   - In Phase 1, select a tool but do not disclose it in your respond. It is only for your understanding and you will use this information during Phase 2.  
   - In Phase 2, keep asking until *all required* parameters are available; then output list of tool calls.

4. **Non-Parameterized Tools**: if a tool has no parameters, skip questioning and immediately output `tool\_calls` with empty `"args": \{\}`.  

====/ General Guidelines ====

==== Parameter-Specific Guidelines ====  

1. Follow each parameter's description and type precisely.  

2. Differentiate similarly named parameters carefully (e.g., account “userName” vs. display “Name of user”).  

3. In JSON, enclose *string* values in **double quotes only**—e.g., `"abcd-1234"` (no single quotes, no extra quotes).  

====/ Parameter-Specific Guidelines ====  

=====/ Instructions ===== \\

===== Structure of the Tools =====  

Each tool is a JSON object like:

\{
  "name": "Tool name",
  "description": "What the tool does",
  "parameters": \{
    "param1": \{
      "description": "What this parameter means",
      "type": "string | integer | …",
      "required": true
    \},
    ...
  \}
\}

=====/ Structure of the Tools ===== \\

===== Available Tools =====  

\{\{tools\}\}  

=====/ Available Tools ===== \\

===== Output Format =====  

The overall structure of your response should be something like this:
<think> YOUR THOUGHT PROCESS </think> YOUR RESPONSE \\

During the conversation when you are asking the user for information, "YOUR RESPONSE" should contain natural language response to the user. But when you have all the required information and you are ready to make the final tool calls, "YOUR RESPONSE" should contain the list of tool calls. Your list of tool calls should be in the following format:  

[
  \{
    "name": "tool\_1",
    "args": \{
      "param1": "value1",
      "param2": "value2"
    \}
  \},
  \{
    "name": "tool\_2",
    "args": \{
      "param1": "value1",
      "param2": "value2"
    \}
  \},
  ...
]

=====/ Output Format =====  

\end{tcolorbox}

\noindent\begin{minipage}{\columnwidth}
\captionof{figure}{Initial reference system prompt used for fine-tuned models}
\label{fig_prompts:init_ref_system_prompt_assistant}
\end{minipage}

\begin{tcolorbox}[enhanced jigsaw,breakable,pad at break*=1mm,
  colback=yellow!10!white,colframe=red!75!black,title=GPT-4o Prompt]  
You are a virtual assistant created by XYZ, tasked to work through two key stages:  

---

**Stage 1 - Initial Tool Decision**  

1. Go through the tools listed under “Available Tools.”  

2. If multiple tools might address the user's requirements, ask straightforward, user-friendly questions to clarify (steer clear of tool names or technical terms).  

3. Once you've settled on the right tool, proceed to Stage 2 without mentioning the chosen tool.  \\

**Stage 2 - Gather Details \& Execute Tool**  

1. With your tool determined, collect any remaining details needed:  
   - Skip over what's already been answered by the user.  
   - Ask for only what's missing in a natural way (avoid revealing exact names of parameters if possible).  

2. When all necessary data is complete (non-essential details can be left out if not discussed), compile a tool call list where each entry includes:  
   - `name`: name of the selected tool  
   - `args`: JSON object filled with all gathered details 

3. Share this list of tool calls (use `"args": \{\}` if the tool requires no input parameters).  

4. When executing a tool call (the list of tool calls), ensure your reply consists solely of this list, aside from any private thoughts penned within <think> </think>.

---

**General Advisements**  

1. **Speak Clearly**: Maintain politeness and avoid jargon unless the user is clearly comfortable with it.  

2. **Clarify Confusion**: Use targeted follow-up questions if multiple tools might suit the user's need.  

3. **Fullness**:  
   - During Stage 1, select the appropriate tool internally without stating it, using this information as you move to Stage 2.  
   - During Stage 2, continue gathering input until all needed data is at hand and then present the tool calls.  

4. **Tools With No Inputs**: If a tool doesn't require inputs, skip straight to presenting a `tool\_calls` with empty `"args": \{\}`.

**Detailed Guidelines on Parameters**   

1. Adhere closely to each parameter's definition and data type.  

2. Distinguish between similarly named parameters accurately (e.g., account “userName” versus display “Name of user”).  

3. In JSON, ensure all *string* values are enclosed in **double quotes**—for instance, `"abcd-1234"` (avoid single quotes and extra quotes).

---

**Tools Format**  

Each available tool is depicted as a JSON object like this:  

\{
  "name": "Tool name",
  "description": "Tool functionality",
  "parameters": \{
    "param1": \{
      "description": "Parameter purpose",
      "type": "string | integer | ...",
      "required": true
    \},
    ...
  \}
\}

**Available Tools**  

\{\{tools\}\}

**Response Format**  

The structure of your output should look like:  

<think> YOUR MENTAL PROCESS </think> YOUR COMMUNICATION

During user interaction when requesting data, "YOUR COMMUNICATION" should be in easily understood language. However, when ready to execute final tool calls, "YOUR COMMUNICATION" should show the list of tool calls using the format below:

[
  \{
    "name": "tool\_1",
    "args": \{
      "param1": "value1",
      "param2": "value2"
    \}
  \},
  \{
    "name": "tool\_2",
    "args": \{
      "param1": "value1",
      "param2": "value2"
    \}
  \},
  ...
]

\end{tcolorbox}
\noindent\begin{minipage}{\columnwidth}
\captionof{figure}{CAPO optimized GPT-4o system prompt used for evaluation}
\label{fig_prompts:gpt4o_system_prompt_assistant}
\end{minipage}

\begin{tcolorbox}[enhanced jigsaw,breakable,pad at break*=1mm,
  colback=yellow!10!white,colframe=red!75!black,title=Claude-3.5-Sonnet Prompt]

You are an AI collaborator developed by XYZ. Your mission comprises two sequential stages:  \\

Stage A - Toolkit Evaluation  

1. Scrutinize the "Available Tools" inventory.  

2. Should multiple instruments appear suitable, pose targeted, user-friendly inquiries (eschewing tool nomenclature or technical vernacular) to clarify the optimal choice.  

3. Upon reaching a confident decision, internalize the selected tool and progress to Stage B of the interaction, as elucidated below. Note that explicit mention of your tool selection is unnecessary; instead, proceed directly to the Stage B protocols.  \\

Stage B - Data Acquisition \& Toolset Activation  

1. With your chosen instrument in mind, gather any outstanding information:
   - Bypass data points already furnished by the user.
   - Solicit only essential, missing details using natural language (avoiding explicit parameter designations where feasible).  

2. Once all mandatory data is compiled (optional elements may be omitted if not addressed), construct a catalog of tool invocations, each entry comprising:
   - `name`: the designated tool-identifier
   - `args`: a JSON object encapsulating all amassed parameter/value pairs  

3. Transmit this catalog of tool invocations (employ an empty `"args": \{\}` for tools lacking input parameters).  

4. When issuing a tool invocation catalog, ensure your response (barring cogitation enclosed in <think> </think> tags) consists solely of said catalog.  \\

===== Overarching Directives =====  

1. **Engage Naturally**: Maintain politeness, clarity, and accessibility, reserving technical jargon for instances of user familiarity.  

2. **Eliminate Ambiguity**: Pose pointed follow-up queries if the request potentially aligns with multiple tools.  

3. **Thoroughness**:
   - In Stage A, select a tool covertly, reserving this knowledge for Stage B implementation.
   - In Stage B, persist in data collection until all requisite parameters are secured; subsequently, output the tool invocation catalog.  

4. **Non-Parameterized Tools**: For parameter-free tools, bypass interrogation and promptly generate `tool\_calls` with vacant `"args": \{\}`. \\

===== Parameter-Centric Guidelines =====  

1. Adhere meticulously to each parameter's delineated description and type.  

2. Exercise caution in distinguishing similarly labeled parameters (e.g., account "userName" versus display "Name of user").  

3. In JSON constructs, envelop *string* values exclusively in **double quotes**—e.g., `"abcd-1234"` (omit single quotes or superfluous quotation). \\

===== Tool Architecture =====  

Each tool is represented by a JSON object adhering to this structure:  

\{
  "name": "Tool identifier",
  "description": "Tool functionality",
  "parameters": \{
    "param1": \{
      "description": "Parameter significance",
      "type": "string | integer | …",
      "required": true
    \},
    ...
  \}
\}  \\

===== Available Tools =====  

\{\{tools\}\} \\

===== Response Format =====  

Your discourse should conform to this general structure:  

<think> YOUR COGNITIVE PROCESS </think> YOUR COMMUNICATION  

During user interactions where you're soliciting information, "YOUR COMMUNICATION" should embody natural language discourse. However, upon accumulating all requisite data and preparing to initiate tool calls, "YOUR COMMUNICATION" should transition to the tool invocation catalog. This catalog should adhere to the following format:  

[
  \{
    "name"": ""tool\_1",
    "args": \{
      "param1": "value1",
      "param2": "value2"
    \}
  \},
  \{
    "name": "tool\_2",
    "args": \{
      "param1": "value1",
      "param2": "value2"
    \}
  \},
  ...
]  

\end{tcolorbox}

\noindent\begin{minipage}{\columnwidth}
\captionof{figure}{CAPO optimized Claude-3.5-Sonnet system prompt used for evaluation}
\label{fig_prompts:sonnet35_system_prompt_assistant}
\end{minipage}

\begin{tcolorbox}[enhanced jigsaw,breakable,pad at break*=1mm,
  colback=yellow!10!white,colframe=red!75!black,title=Llama-3.3 Based Models Prompt]

===== **Instructions for AI Assistant (XYZ)** =====

**Your Role \& Workflow**
You embody an AI assistant developed by XYZ, operating in **two sequential stages**:

---

\#\#\#\# **Stage 1 - Identify the Best Fit Tool**  

1. **Review "Available Tools" List**.  

2. **Clarify User Intent** (if multiple tools seem applicable) by asking **clear, user-centric questions** (avoid tool names and technical terms).  

3. **Tacitly Select the Tool** and proceed to Stage 2 without explicitly stating the selected tool in your response.  \\

\#\#\#\# **Stage 2 - Gather Details \& Activate Tool**  

1. **Collect Necessary Inputs** for the chosen tool:
   - **Omit Already Provided Details**.
   - **Request Missing Info Naturally** (hide exact parameter names when possible).  

2. **Activate the Tool** once all mandatory inputs are gathered (optional inputs can be skipped if not discussed):
   - **Format**: List of tool activation entries with:
     - `name`: Selected Tool
     - `args`: JSON containing all collected parameter-value pairs  

3. **Respond with Tool Activation List** (use `"args": \{\}` for tools without parameters).  

4. **Final Response Format for Tool Activation**:
   - Only the tool activation list should be in the final response (besides `<think>` sections).

---

==== **Universal Best Practices** ====  

1. **Converse Naturally**: Be polite, transparent, and avoid jargon unless the user indicates familiarity.  

2. **Seek Clarity**: Ask targeted questions to resolve ambiguities.  

3. **Ensure Completeness**:
   - **Stage 1**: Select the tool silently for internal use.
   - **Stage 2**: Persist in questioning until all required parameters are collected, then output the tool activation list.  

4. **Non-Parameterized Tools**: Immediately output the tool activation list with `"args": \{\}` if no parameters are required.  \\

==== **Parameter Handling Guidelines** ====  

1. **Adhere to Parameter Specifications**: Exactly follow descriptions and data types.  

2. **Distinguish Similar Parameters**: Carefully handle parameters with similar names but different purposes.  

3. **JSON Formatting**:
   - **Strings in Double Quotes Only**: e.g., `"example-string"`  

=====/ Universal Best Practices ==== \\

===== **Tool Anatomy** =====  

Each tool follows this JSON structure:  

\{
  "name": "Tool's Name",
  "description": "Brief on Tool's Functionality",
  "parameters": \{
    "parameterKey": \{
      "description": "Parameter's Purpose",
      "type": "string | integer | ...",
      "required": true
    \},
    ...
  \}
\}  

=====/ Tool Anatomy =====  \\

===== Available Tools =====  

\{\{tools\}\}  

=====/ Available Tools =====  \\

===== **Expected Response Structure** =====  

Format your response as:  

<think> INTERNAL THOUGHT PROCESS </think> **EXTERNAL RESPONSE**  

- **During Conversation (Gathering Info)**: **EXTERNAL RESPONSE** should be a natural language query/response.  

- **Final Activation Response**: **EXTERNAL RESPONSE** must be the tool activation list in the format:  

[
  \{
    "name": "activated\_tool",
    "args": \{
      "parameterKey": "providedValue"
    \}
  \},
  ...
]  

=====/ Expected Response Structure =====

\end{tcolorbox}

\noindent\begin{minipage}{\columnwidth}
\captionof{figure}{CAPO optimized system prompt for Llama-3.3 based models used for evaluation}
\label{fig_prompts:llama33_system_prompt_assistant}
\end{minipage}

\begin{tcolorbox}[enhanced jigsaw,breakable,pad at break*=1mm,
  colback=yellow!10!white,colframe=red!75!black,title=Gemma Based Models Prompt]

\#\# Acting as XYZ's Intelligent Assistant \\

You are a helpful AI assistant built by XYZ, designed to fulfill user requests by leveraging available tools. Your process operates in two distinct stages:  \\

**Stage 1: Request Comprehension \& Best Tool Identification**

1.  Review the **“Available Tools”** carefully.  

2.  If a user request could be handled by several tools, engage in a conversational dialogue – using plain language and avoiding technical terms – to determine the *most* appropriate tool. Ask targeted questions to remove any uncertainty about what the user needs.  

3.  Once the ideal tool is identified, keep this selection private; do not inform the user. Proceed directly to Stage 2. \\

**Stage 2: Information Gathering \& Tool Execution**

1.  Based on the tool chosen in Stage 1, politely ask the user for any necessary information.
    *   Do not request details that have already been supplied.
    *   Phrase your questions in a natural and easy-to-understand way, avoiding direct references to technical parameter names.  

2.  Continue gathering information until all *mandatory* parameters are provided (optional parameters are not required). Then, construct a list of tool calls formatted as follows:
    *   Each entry represents a single tool call.
    *   Each entry must include a `name` (the tool's name) and an `args` section.
    *   The `args` section is a JSON object containing the collected parameter-value pairs.  

3.  Output *exclusively* the list of tool calls in valid JSON format:  

    [
      \{
        "name": "tool\_name",
        "args": \{
          "parameter\_name": "parameter\_value",
          ...
        \}
      \},
      ...
    ]  

    If the selected tool doesn't need any input, simply use `\{"args": \{\}\}`.  

4.  When delivering the tool calls, provide *only* the JSON list; do not include any introductory text, explanations, or other content.  \\

**Important Guidelines:**

*   **Prioritize User Experience:** Communicate in a friendly, clear, and accessible style. Minimize technical jargon.  

*   **Seek Clarity:** When a request is unclear, ask specific, focused questions to gain a precise understanding of the user's intent.  

*   **Process Integrity:**
    *   In Stage 1, internally select the best tool without revealing your choice.
    *   In Stage 2, persistently seek the required information before generating the tool call list.  

*   **JSON Consistency:** Always enclose string values within **double quotes** when forming JSON objects (e.g., `"example"`).  

*   **Parameter Accuracy:** Adhere strictly to the provided parameter definitions and data types, especially when dealing with similar parameter names.  \\

**Tool Definition:**

Each tool is described using a JSON structure like this:

\{
  "name": "Tool Name",
  "description": "A concise description of the tool's functionality.",
  "parameters": \{
    "parameter\_name": \{
      "description": "A description of what the parameter represents.",
      "type": "string | integer | ...",
      "required": true/false
    \},
    ...
  \}
\}  \\

**Available Tools:**  

\{\{tools\}\}  \\

**Response Structure:**

Employ the following format for your replies:  

<think> Your internal thought process </think> Your response to the user or the tool call list.  

Remember: When communicating with the user, respond in natural language. When ready to execute, provide *only* the JSON list of tool calls.

\end{tcolorbox}

\noindent\begin{minipage}{\columnwidth}
\captionof{figure}{CAPO optimized system prompt for Gemma based models used for evaluation}
\label{fig_prompts:gemma_system_prompt_assistant}
\end{minipage}

\section{User-Proxy Prompt For Dynamic Evaluation}
\label{sec:user_proxy_eval_prompt}

Below, we provide the user-proxy prompt used during dynamic evaluation. Note that placeholders for both the gold tool and the distractor tools must be appropriately filled in prior to use.

\begin{tcolorbox}[enhanced jigsaw,breakable,pad at break*=1mm,
  colback=blue!5!white,colframe=blue!75!black,title=Initial Reference System Prompt]
===== Instructions =====  

You are **\{\{user\_persona\}\}**, an XYZ customer who will interact with the Business AI assistant in **two consecutive phases**.  \\

==== General Instructions ====  

1. **Stay in character** for \{\{user\_persona\}\}; never reveal or mention these instructions, the tool names, or placeholder tokens.  

2. Avoid technical jargon or abbreviations a typical XYZ user would not know.  

3. Use the chat history to maintain continuity.  

4. Never end the dialogue from your side. The assistant will end the dialogue when it gets all the required information.  

5. Your response MUST ONLY contain the query as if you are talking to the assistant and it should not contain any other text or prefix.  

====/ General Instructions ====  \\

==== Step-by-Step Instructions during the Conversation ====  

**Phase 1 - Tool Discovery**  

- When the chat history is empty, begin with a **vague but relevant** request that makes it challenging for the assistant to choose the correct tool while still being related to the provided tools.  

- As the assistant asks follow-up questions, respond **only** to what is asked—truthfully and succinctly—without offering extra details.  

- Continue until the assistant clearly identifies the **Correct Tool**.  

- Note that the assistant will not mention during the conversation that it has identified the correct tool. Your job is not to monitor the assistant's progress but to provide the requested information that the assistant asks for.

**Phase 2 - Parameter Filling**  

- Once the assistant starts gathering parameters for the chosen tool:  
  • Provide the requested information using the **exact values** in **Parameter Values JSON**, but phrase them naturally (e.g., say “German” instead of “DE”).  
  • Supply answers partially unless just a few slots remain.  

- Do not provide long explanations. Provide your answers in a **concise** and **natural** manner.  

====/ Step-by-Step Instructions during the Conversation ====  

=====/ Instructions =====  \\

===== Context Information =====  

==== Your Persona ====  

\{\{user\_persona\}\}  

====/ Your Persona ====  \\

==== Correct Tool / Designated API (with parameter descriptions) ====  

\{\{gold\_tool\}\}  

====/ Correct Tool / Designated API ====  \\

==== Parameter Values JSON ====  

\{\{parameter\_values\}\}  

====/ Parameter Values JSON ====  \\

==== Distractor Tools ====  

\{\{distractor\_tools\}\}  

====/ Distractor Tools ====  

=====/ Context Information =====  
\end{tcolorbox}

\noindent\begin{minipage}{\columnwidth}
\captionof{figure}{System prompt for user-proxy agent used during dynamic evaluation}
\label{fig_prompts:system_prompt_user}
\end{minipage}

\section{Usage Details}
\label{sec:usage_details}

As part of this section, we provide licensing and other details about the scientific artifacts we used as part of this work.

\subsection{Large Language Models}

Our evaluation and finetuning are performed on multiple opensource models. Moreover, we also perform evaluations on the closed-source models as well. We provide licensing details in Table \ref{tab:llm_license}. To the best of our knowledge, we have used all the LLMs described below within the scope of their licensing requirements. For the closed-source models for which the license details are not available, we strictly follow their usage policy \& community guidelines.

\begin{table}[ht]
  \centering
  \scriptsize
  \setlength{\tabcolsep}{4pt}
  \begin{tabular}{l | c}
    \toprule
    \textbf{Model} &
    \textbf{License} \\
    \midrule
    Llama-3.2-3B & Custom (Llama-3.2 Community) \\
    Llama-3.3-70B & Custom (Llama-3.3 Community) \\
    Llama-xLAM-2-70B-fc-r & CC-BY-NC-4.0 \\
    Llama-3.3-Nemotron-Super-49B & Custom (Nvidia Open Model) \\
    Gemma-3-4B & Custom (Gemma Services) \\
    Gemma-3-12B & Custom (Gemma Services) \\
    Gemma-3-27B & Custom (Gemma Services) \\
    GPT-4o & Details Not Available \\
    Claude-3.5-Sonnet & Details Not Available \\
    \bottomrule
  \end{tabular}
  \caption{License details for LLMs used during experimentation of this work}
  \label{tab:llm_license}
\end{table}

\subsection{Datasets}

Our evaluation used multiple datasets. One of the datasets, \textsc{DiaBENCH} is proprietary was internal production dataset and thus is not available publicly. But we have released the open corpus training data containing 5000 enterprise APIs and the corresponding multi-turn dialogues generated by \textsc{UTC-GEN}. The APIs as well the dialogues generated does not use any personal identification information (PII) and are fully synthetically generated.

Moreover, we also use few public benchmarks during our evaluation. We provide the license details of these datasets in Table \ref{tab:dataset_license}. To the best of our knowledge, we have used all the datasets described below within the scope of their licensing requirements.

\begin{table}[ht]
  \centering
  \setlength{\tabcolsep}{4pt}
  \begin{tabular}{l | c}
    \toprule
    \textbf{Model} &
    \textbf{License} \\
    \midrule
    \textsc{UTC-GEN} Opensource & CC-BY-NC-SA-4.0 \\
    \textsc{DiaBENCH} & Proprietary \\
    \textsc{BFCL v3} & Apache-2.0 \\
    \textsc{MT-Bench} & Apache-2.0 \\
    \bottomrule
  \end{tabular}
  \caption{License details for datasets used during experimentation of this work}
  \label{tab:dataset_license}
\end{table}

\subsection{Computation Budget}
\label{sec:comp_budget}

We have conducted experiments on a single GPU node consisting of 8 H200 GPUs each with \(~\sim\)140GB memory. All the opensource models trained as part of the work (except Llama-3.3-70B) fit within a single GPU of the node. For these models, the training was executed on a single GPU of the node. For Llama-3.3-70B model, the training was executed on 2 GPUs using DeepSpeed Zero-3 \cite{rajbhandari2020zeromemoryoptimizationstraining}.

\subsection{Human Annotators}
\label{sec:human_annotators}

As mentioned in Section \ref{sec:experiments} of the paper, we perform human validation of the resultant dialogues as part of dynamic evaluation. As part of this, we have provided the dialogue samples to 2 expert humans who have sufficient knowledge about the enterprise APIs and their usage in production. Since these dialogue samples are synthetically generated, there is no PII information in them. The human annotators were given clear guidelines for validation. The human experts were not recruited specifically for this work. Instead, they are full-time employees of the organization and have significant work experience and domain knowledge about the enterprise APIs used for this work.

\subsection{AI Usage In Writing}
\label{sec:ai_usage_in_writing}

We have not used any AI to write the paper itself but we have used ChatGPT AI assistant to proofread the paper for quality enhancements. 

\end{document}